\documentclass[final,5p,times,twocolumn]{elsarticle}

\usepackage{amssymb}
\usepackage{amsmath}
\usepackage{multirow}
\usepackage{lineno}
\usepackage{url}
\newcommand{\R}{\mathbb{R}}
\usepackage{graphicx}
\usepackage{subfigure}

\journal{Automation in Construction}

\begin{document}
\begin{frontmatter}



\title{Deep Learning for Segmentation of Cracks in High-Resolution Images of Steel Bridges}

\author[label1,label2]{Andrii Kompanets}
\author[label3]{Gautam Pai}
\author[label2,label3]{Remco Duits}
\author[label1,label2]{Davide Leonetti}
\author[label1]{H.H. (Bert) Snijder}

\affiliation[label1]{organization={Eindhoven University of Technology, Department of the Built Environment},
            city={Eindhoven},
            country={the Netherlands}}

\affiliation[label2]{organization={ Eindhoven Artificial Intelligence Systems Institute, Eindhoven University of Technology},
            city={Eindhoven},
            country={the Netherlands}}

\affiliation[label3]{organization={Eindhoven University of Technology, Department of Mathematics and Computer Science},
            city={Eindhoven},
            country={the Netherlands}}



\begin{abstract}
Automating the current bridge visual inspection practices using drones and image processing techniques is a prominent way to make these inspections more effective, robust, and less expensive. In this paper, we investigate the development of a novel deep-learning method for the detection of fatigue cracks in high-resolution images of steel bridges. First, we present a novel and challenging dataset comprising of images of cracks in steel bridges. Secondly, we integrate the ConvNext neural network with a previous state-of-the-art encoder-decoder network for crack segmentation. We study and report, the effects of the use of background patches on the network performance when applied to high-resolution images of cracks in steel bridges. Finally, we introduce a loss function that allows the use of more background patches for the training process, which yields a significant reduction in false positive rates.
\end{abstract}




\end{frontmatter}



\section{Introduction}
The structural integrity of a bridge is affected by various factors such as fatigue, corrosion, possible traffic collisions, and even natural disasters. These phenomena and events can affect the structural integrity of the bridge and lead to a catastrophic collapse  \cite{zhang2022causes,garg2022analysis,Imam210MetallicBridgeFailureStatistics}. 
Furthermore, recent statistics show the need for careful attention to the condition and the safety of bridges.  
As of 2023, in the USA out of 621,510 bridges, 6.8\% are considered structurally deficient and 36\% require major repair work, according to a report in \cite{ARTBA}. 
Meanwhile, in Europe, bridges have aged considerably \cite{brady2009cost345}.
For example, in the Netherlands, about 80\% of steel bridges and about 20\% of concrete bridges have a remaining design life in the range between 0\% and 33\% according to a bridge condition assessment made in~\cite{Rijkswaterstaat2022ReportCondition}. 

Bridge maintenance is carried out as a measure to prevent degradation processes from reaching a critical stage, which can potentially lead to unexpected collapse. 
Within maintenance, bridge inspection aims at detecting damage, and its results
 are used for both: assessing the structural integrity of the bridge as well as planning
further maintenance steps, such as remedial works or subsequent inspections. 

Based on local regulations, there are different requirements for inspection procedures and techniques.   
According to \cite{brady2009cost345}, in European countries and in the USA inspection scope can range from superficial or routine inspections to major or in-depth inspections.
Routine inspections are performed to identify extensive damage and asses the general state of the structure.
In contrast, major inspections involve a close examination of all accessible bridge parts. 

A range of inspection techniques can be used during the major inspections.
Thanks to its simplicity, visual inspection is the most used technique for the inspections of bridges even though it is capable of detecting only surface defects of relatively large size, and its performance is significantly affected by aspects such as subjective assessment by individuals or unpredictable weather conditions \cite{campbell2020PODVisual,nepomuceno2022RoboticsSurvey}.  On the contrary, advanced Non-Destructive Testing (NDT) techniques such as ultrasonic testing are capable of detecting relatively small defects even below the inspected surfaces \cite{hopwood2016nondestructive}. However, due to their significant effort, cost, and complexity, advanced NDT techniques are mostly used for closer examination of damage that was initially detected by visual inspection, or for inspection of fracture-critical parts of the bridge structure \cite{hopwood2016nondestructive}.

As visual inspection is the most frequent type of inspection, in this paper we make an effort to augment current visual inspection practices by developing an algorithm for automatic crack detection on images. 

To conduct a regular visual inspection of a bridge structure, a trained inspector visually observes the bridge structure assessing its condition. 
Moreover, inspectors can use magnifying optics or tools to clean the inspected surface or locally remove paint when this is suspected to cover a crack. 
Special vehicles such as bucket trucks or climbing equipment and ropes can be used to gain outstretched hand access to different parts of the bridge  \cite{lovelace2015UAVInspectionDemonstrationProject}. 
However, gaining close access to highly elevated structures can raise safety concerns for both inspectors and the public.  
Additionally, the necessity to work at a high altitude near the bridge may affect bridge traffic and may require additional training for inspectors \cite{khedmatgozar2021NonDestractiveTesting}. 

With advancements in robotic technologies, the aforementioned drawbacks of visual inspections can be eliminated. 
Numerous studies show the possibility and benefits of applying robots for bridge inspection, which are reviewed in \cite{nepomuceno2022RoboticsSurvey,ahmed2020ReviewRoboticInspection}. 
Ground robots and climbing robots are convenient for automating contact NDT such as ultrasonic inspection, since they are often located closely or even mounted to the inspected surface.  
In contrast, aerial robots or Unmanned Aerial Vehicles (UAVs)  have much higher mobility and are capable of reaching inspected surfaces relatively easily and fast. However, for the safe operation of the UAVs they have to operate from a distance to the inspected surface to prevent collision unless it is specifically designed for this purpose, as done in \cite{ikeda2017WallContactUAV,tognon2019AerialManipulatorSystemInspectionIndustrialPlants}. 
Therefore, the use of UAVs is intended as an advancement to improve current visual inspection practices by collecting images of bridge details to be inspected \cite{ahmed2020ReviewRoboticInspection}.
When used as a tool for visual inspection, UAVs have a great potential to reduce inspection cost by up to 75\% and time by up to 80\%, as explained in \cite{balagopalan2018SimulationDroneInspectionCost}. 
Furthermore, 
the reliability of UAV-aided inspection can reach and even overcome the reliability of conventional visual bridge inspection \cite{kim2022CompareManualandUASInspection}.
However, long image post-processing time can be one of the drawbacks of this approach, which limits the applicability of UAVs for visual inspection \cite{jongerius2018RijkswatersaatUAVInspection} since it is common to have a delay due to a human inspector analyzing these images which in some cases can be as long as a couple of weeks. 

Alternatively, a computer vision algorithm can be used to reduce the time required for the analysis of the images, thus automating the inspection procedure.
Apart from a gain in speed and cost, automatic damage detection on images collected by a UAV can also be more accurate and robust. As demonstrated in \cite{kim2022CompareManualandUASInspection}, automatic inspection enabled the detection of cracks that were undetected in a  conventional visual bridge inspection. 
Apart from automatic visual inspection, the same algorithms can be also used in conjunction with augmented reality and digital twin technologies as was proposed in \cite{ccelik2022BridgeInspectionAugumentedReality}. Therefore, in this paper, we intend to present a novel automatic fatigue crack detection algorithm on the images that were collected during inspection. 

\subsection{Deep learning for crack detection}

Different computer vision algorithms for the automatic detection and measurement of structural damage in bridges have been developed over the past years ranging from classical image processing to data-dependent machine learning models. 
Recent advancements in the field of deep learning and convolutional neural networks (CNNs) have enabled impactful advancements in automatic structural damage detection. 
Since fatigue cracks are one of the major concerns during visual inspections \cite{hopwood2016nondestructive}, crack segmentation and detection with deep learning has gained significant interest among researchers, leading to many methodological contributions as enumerated comprehensively in the literature reviews: 
\cite{hamishebahar2022comprehensivereview,ali2022bibliometricreview}.

Three main computer vision methods are used for the identification of cracks on images, namely: \textit{image classification}, \textit{object detection}, and \textit{image segmentation} \cite{hamishebahar2022ReviewDeepLearningCrackDetection}. 
\textit{Image classification} algorithms can indicate if there are cracks captured on the image and provide a classification of its type \cite{Li2020CrackClassification}. 
\textit{Object detection} algorithms extend the image classification by giving an approximate location of the cracks on the image using a so-called bounding box \cite{li2023automatic}. 
With \textit{image segmentation} algorithms it is possible to get very accurate crack information since each pixel in the inspected image is classified by the automatic segmentation algorithm as a crack pixel or a background pixel.  Naturally, this gives an accurate location of the crack on the image and allows for the measurement of its geometry (length, width, etc.) \cite{fan2022use}. 
The importance of crack measurement for the bridge maintenance workflow was indicated in \cite{jongerius2018RijkswatersaatUAVInspection}. 
Therefore,  the \textit{crack segmentation} approach invokes most of the research interest.

A sophisticated review of the publications specifically dedicated to crack segmentation is reported in \cite{li2022review}. 
Neural networks based on U-Net architecture \cite{ronneberger2015unet} or encoder-decoder networks are the most commonly used for crack segmentation. 
Numerous publications explore the effect of different choices on the performance of the encoder-decoder network for crack segmentation, namely: (1) neural network architecture \cite{islam2019crackSegnet,zou2018deepcrack,yang2021automatic,miao2019automatic,yang2022nondestructive,qu2021crack,sun2020roadway,chen2021HACNet,jiang2020HDCB-Net}; (2) loss function \cite{cheng2018pixel,Takumi2019f1basedOptimisationCrackDetection}; (3) post-processing \cite{zhang2019concrete}; (4) transfer learning \cite{yang2021automatic,ccelik2022sigmoid,lau2020automated}. It was shown that the encoder-decoder networks can be effective for the crack segmentation task, especially when used with the Dice loss function \cite{sudre2017DICE_loss} combined with binary cross entropy loss \cite{cheng2018pixel} and when used with in-domain or cross-domain transfer learning.

Generative Adversarial Network (GAN) based approaches were less used for crack segmentation. 
Unlike other approaches, GAN-based methods can perform crack segmentation on a set of images that are not annotated, i.e. unsupervised learning \cite{duan2020unsupervisedCrackSegmentation,zhang2020self,li2020semi}, or with partially accurate annotation or a small number of annotated images, i.e semi-supervised learning \cite{li2020semi,zhang2020crackgan}. 
However, the performance of such methods is usually lower as compared to the supervised learning methods. 

Transformer networks \cite{liu2021SwinTransformer,dosovitskiy2020ViT} are trending in the field of computer vision, and when introduced they were showing superior performances on large datasets over CNNs. CNNs, rely on inductive biases and geometrical priors such as local correlations, feature hierarchy, and translational equivariance. The novel idea of the transformer networks lies in rejecting most of the geometric priors and relying on massive datasets instead. 
For example, the transformer-based Segment Anything model \cite{kirillov2023SegmentAnythingModel} was pre-trained on a dataset with 300 million images and fine-tuned on a dataset containing 1 million images.
Pure transformer networks are rarely used for crack segmentation \cite{guo2023PureTransformerforCracks} and often a combination of transformers with CNNs is applied \cite{quan2023crackvit,wang2022automatic,xie2021segformer}. However, a recently introduced CNN called ConvNext \cite{Liu_2022ConvNext} showed that the classical CNN can have better performance than the transformer networks while being easier to implement and less demanding on training \cite{Liu_2022ConvNext}.  

Another class of methods is Geometric Deep Learning. In comparison to transformers and CNNs, geometric learning methods employ more inductive bias and geometric priors in the neural architectures. Such methods are known for good performance in scenarios of limited availability of training data and have inherent roto-translation equivariance. Moreover, due to the reduced number of trainable parameters, geometric learning methods also improve the transferability to other datasets. Such data-efficiency and transferability properties of roto-translation equivariance geometric learning on small/mid-size labeled datasets are also reported on line segmentation tasks in \cite{pai2023}. An example of the application of such geometric learning methods to crack segmentation is ARF-Crack \cite{chen2020ARFCrack} where the neural network performance is enhanced by using active rotation filters (ARF) \cite{Zhou2017ARF}. 

Roto-translation equivariance from geometric learning methods is generally desirable by design. However, potent architectures like ConvNext together with large annotated datasets and data augmentations allow for learning these geometric invariances from data alone. Furthermore, CNN architectures have a strong local pixel adaptivity guided by the training data, whereas geometric learning relies on intuitive geometric priors \cite{bellaard2023analysis} that can disturb favorable local pixel adaptivity. 

In the data collection and processing stage, we used effective geometric semi-automatic annotation tools \cite{CP1} to generate accurate ground truth labels for the cracks (see Appendix~\ref{app:A}). This allowed us to easily generate a large amount of well-annotated training data that is important for the impactful use of ConvNext architectures. As a result, we could avoid the use of computationally more demanding geometric deep learning architectures. 

In summary, we integrate all these considerations and promote the design of an architecture that combines ConvNext \cite{Liu_2022ConvNext} with an encoder-decoder approach proposed by \cite{Konig2021OED}. Indeed, as reported in Section \ref{sec:Results and discussion}, our method shows a competitive performance on the large, annotated dataset of cracks in steel bridges that we introduce. 




\subsection{Crack Detection in Steel Bridges}

Most of the research concerning crack detection deals with images of cracks in road pavement and concrete, whilst crack detection in steel structures has been investigated less, probably due to the lack of open-access datasets.  
A dataset with images of cracks in steel bridge girders was presented in \cite{bao2021SimilarDataset,xu2019SimilarDataset2}. 
It was not made publicly available and only researchers who participated in the competition IPC-SHM 2020   \cite{bao2021SimilarDataset} were given access to it.  
The database consists of 120 pixel-wise labeled images with sizes of 4928×3264 and 5152×3864 pixels. Additionally, 80 un-labelled images were collected which are meant to be used for testing and validation. 
The images were collected by hand-held camera from various distances to the bridge structure. 
Often, the images contain a steel surface with a crack developing at a welded connection, where handwritten markings are also present, thus complicating the segmentation task. 

Since in our work, we deal with similar images, in this subsection, we take a closer look at the publications that present algorithms for the detection and segmentation of cracks in the images of cracks from the IPC-SHM 2020 dataset \cite{bao2021SimilarDataset,xu2019SimilarDataset2}. 
These publications are summarised in Table \ref{tab:steel cracks segmentation}.  

\setlength{\tabcolsep}{2pt} 
\begin{table}
\caption{Methods for crack detection in steel bridges based on the similar but smaller dataset from IPC-SHM 2020 \cite{bao2021SimilarDataset} \label{tab:steel cracks segmentation}. For the methods tested on our dataset we get similar performance values, cf.~Table~\ref{tab:Results on CSB}. The $Pr$, $Re$ and $F1$ metrics are explained in Section \ref{sec:Experiments:Evaluation}}

\centering
\begin{scriptsize}
\begin{tabular}{cp{2cm}p{33pt}l} 
\hline    
Method & Approach & \multicolumn{2}{c}{Performance}\\ 
\hline 
Xu et. al. \cite{xu2019SimilarDataset2}& Patch classification & Crack:  &$Pr$=93.9\% $Re$=95.5\% $F1$=94.7\%
\\ 
 &   & Handwriting: &$P$r=97.6\% $Re$=97.2\%  $F1$=97.4\%
\\ 
 &   & Background:  &$P$r=95.3\% $Re$= 94.0\% $F1$=94.6\% 
\\
 & & \multicolumn{2}{c}{}\\ 
Quqa et. al. \cite{quqa2022two} & Patch classification  & \multicolumn{2}{l}{$Pr$ = 95.5\% $Re$=93.9\%, $F1$=94.7\%}\\ 
 & \& postprocessing  &  \multicolumn{2}{c}{}\\
 & & \multicolumn{2}{c}{}\\ 
Tong et.al. \cite{tong2021SteelDatasetFeedbackUpdate}& Patch classification & \multicolumn{2}{l}{IoU = 53.56\%}\\ 
 & \& segmentation  &  \multicolumn{2}{c}{}\\
 & & \multicolumn{2}{c}{}\\ 
Dong et. al. \cite{dong2021SteelDatasetUNet}& Segmentation & \multicolumn{2}{l}{IoU = 65.06\%, $F1$=42.95\%}\\
 & & \multicolumn{2}{c}{}\\  
Zhang et.al. \cite{Zhang2022SteelDatasetEnsemble3}& Patch classification & \multicolumn{2}{l}{$Pr$=80.76\% $Re$=81.76 IoU=67.99\%}\\
 & \& segmentation  &  \multicolumn{2}{c}{}\\
 & & \multicolumn{2}{c}{}\\ 
Han et.al. \cite{han2022detection}& Detection & \multicolumn{2}{l}{IoU=61.34\%}\\ 
 & \& segmentation  &  \multicolumn{2}{c}{}\\
 & & \multicolumn{2}{c}{}\\ 
Meng et.al. \cite{meng2023real}& Image classification & \multicolumn{2}{l}{$Pr$=88.07\% $Re$=88.30\% $F1$=78.69\%}\\ 
 & \& segmentation  &  \multicolumn{2}{c}{}\\ \hline

\end{tabular}
\end{scriptsize}
\end{table}

In \cite{xu2019SimilarDataset2} a patch classification approach is used to detect cracks. 
A CNN is used to classify image patches into three categories: cracks, background, and handwriting. The classification of handwriting patches is done to decrease the number of false positive crack detections because the handwriting can easily be misrecognized by a neural network as a crack. 
Moreover, this work aimed at improving crack detection results by using super-resolution images. 
To increase the resolution of images and to make cracks more recognizable, a super-resolution neural network is used. 
Surprisingly, when the neural network for crack detection is trained and evaluated on the images with increased resolution, the crack detection performance decreases. 
These results can be explained by the fact that the super-resolution neural network is a generative network, which means that it produces realistic images that can not necessarily be an accurate representation of an actual object that was photographed. 

In \cite{quqa2022two} the patch classification approach is extended with a subsequent post-processing step to refine the results of patch classification using the information about the mutual arrangement of patches where a crack is identified. 
Furthermore, classical image processing techniques are used to measure the length and width of the detected cracks.

Often, images of steel bridge structures contain several features similar in appearance to cracks which can be misrecognized as cracks, such as handwritten markings, shadows, and joints of the structure. Throughout the rest of the paper, we will call these features of the image that look like a crack but are not a crack - crack-like features.
To address the problem caused by these image crack-like features, in \cite{tong2021SteelDatasetFeedbackUpdate} a feedback update training strategy is proposed, which consists of a few training stages, to train a CNN to classify image patches. At the first training stage, random crack patches and random background patches are used to train a neural network for patch classification. 
At the second training stage, background patches for training are taken, in which the trained neural network tends to make a false positive error by classifying the background patch as a crack patch. 
The third and fourth training stages repeat the procedure. 
Furthermore, after the patches of images with cracks are identified, an encoder-decoder segmentation network is applied to segment cracks on those patches. An Intersection over Union (IoU) metric is used to estimate the performance of the neural network, which is reported at the level of 53.56\%. 


An encoder-decoder network is applied to segment entire crack images in \cite{dong2021SteelDatasetUNet}. To train the neural network in \cite{dong2021SteelDatasetUNet} 480 crack patches of size 512x512 were generated. 
Interestingly, it was also shown that sequential application of patch classification and segmentation neural networks can significantly improve crack segmentation results in images with complex backgrounds. 
This is because a classification neural network has a wider field of view and has a better capability to distinguish cracks from crack-like image features \cite{Zhang2022SteelDatasetEnsemble3}. 
In \cite{Zhang2022SteelDatasetEnsemble3} only 35.10\% of IoU is achieved using only a segmentation neural network. 
The performance is almost doubled achieving 63.49\% by using a segmentation neural network to only those image patches that are classified as crack patches by a classification neural network.
Finally, segmentation performance is further improved up to 67.99\% of the IoU, by combining the patch classification and segmentation neural network into a complex pipeline. 

Similarly, in \cite{han2022detection} a two-step approach for crack segmentation is used. 
First, a YOLOv3 algorithm is used to identify the approximate location of cracks while an ensemble segmentation neural network is applied to segment cracks in that region. It is reported that the YOLOv3 algorithm failed to distinguish crack-like features from cracks when the input patch size is 256x256, so the input patch size is increased to 608x608 pixels. 
To segment cracks in the area of the image, which is identified by the YOLOv3 algorithm as crack area, 15 DeepLabv3+ \cite{chen2018DeepLabv3P} encoder-decoder networks are trained on different datasets with different complexity. 
The IoU was reported at the level of 61.34 \%.

Finally, a comprehensive study was made in \cite{meng2023real}. 
There, a BiSeNet V2 \cite{yu2021bisenet} is used as an initial lightweight and fast segmentation neural network for which the results are enhanced with a high-precision crack segmentation algorithm based on U-Net with additional attention layers. 
The overall performance of the combination of these two neural networks is reported at the level of 78.69\%
The work also considers a drone control algorithm, where the flight path of the drone is modified to get closer to a crack, once the lightweight neural network detects a crack.
Further, the physical crack width measured in millimeters is estimated using a width measurement algorithm and a depth map provided by a ZED2 stereo camera. 
All components are integrated into a single inspection system including a drone with a specifically designed control system, crack detection and segmentation algorithms, and a crack physical width measurement algorithm. 
The developed inspection system was tested in real-world conditions, proving the feasibility of the concept of fully automatic visual inspection using a UAV.  

The earlier works show that the combination of a few neural networks into a single crack segmentation system can be beneficial in terms of segmentation accuracy. For example, a combination of a crack detection neural network, which identifies the approximate location of a crack, and a segmentation neural network which segments the crack in that specific region, can be superior to a single segmentation neural network which is applied to the entire image. However, in our work, we focus on the development of a single neural network for crack segmentation on entire images of steel bridges. We left the integration of this neural network into a more complex and robust system for future research.  

\subsection{Contributions of the Article}

There are three main components of the development of a deep learning algorithm: data, model, and training. Our contribution concerns all three components in terms of the development of a deep learning algorithm for crack segmentation. 

Numerous publicly available datasets for the segmentation of cracks in concrete structures or asphalt pavement are available \cite{bianchi2022StructuralInspectionDatasets}, which are listed in Table \ref{tab:Datasets list} of \ref{app:B}. 
However, the detection of cracks in steel has its own peculiarities, and despite the growing scientific interest in the field of automatic visual crack detection \cite{hamishebahar2022ReviewDeepLearningCrackDetection}, up to the best knowledge of the authors, there is no publicly available dataset with images of cracks in steel bridges. 
To close this gap, we collected images of cracks in steel bridges and annotated them for segmentation, using the semi-automatic tool \cite{CP1}. 
The dataset is made publicly available, see \cite{CSB_dataset}

Further, we adopt a recent ConvNext neural network \cite{Liu_2022ConvNext} for the crack segmentation task. This is done by using the ConvNext as an encoder, in the modified encoder-decoder network which was proposed by Konig et al. \cite{Konig2021OED} for crack segmentation and which showed state-of-the-art performance on a few datasets \cite{Konig2021OED}.

Moreover, we explore ways to train a neural network using patches, in a way that allows to achieve the best possible performance on entire high-resolution images. We experiment with different amounts of background patches which are used alongside crack patches to train a neural network. These experiments allowed to highlight the importance of using the right amount of background patches for training a neural network for crack segmentation on entire high-resolution images. 
Previous works on the segmentation of cracks in steel bridges overlook this point, using for training only crack patches \cite{Zhang2022SteelDatasetEnsemble3,Chuanzhi2021SteelUnet}, equal number of background patches and crack patches \cite{tong2021SteelDatasetFeedbackUpdate} or downsizing the images instead of using patches \cite{han2022detection,meng2023real}. 

Based on the results of the study on the effect of background patches, we propose a modification to a commonly used loss function, thus presenting the loss function with inversion, which is intended to allow to use more background patches for the neural network training.  
Finally, we studied the effect of patch size on neural network performance and how it affects the ability of a neural network to distinguish cracks from crack-like image features.

To summarise our main contributions we list them as follows:
\begin{itemize}
\item We introduce a public dataset of images of cracks in steel bridges, annotated for segmentation, see \cite{CSB_dataset}.
\item We adopt a state-of-the-art deep learning neural network namely ConvNext \cite{Liu_2022ConvNext} for the crack segmentation task, combining it with the encoder-decoder network which was designed for crack segmentation \cite{Konig2021OED}.
\item We investigate how to train a neural network to segment high-resolution images using a patch-based approach, by 
optimizing the amount of background patches used for training. 
\item  We propose a modification to a loss function that allows the use of more background patches, which is crucial for optimal performance on entire high-resolution images.
\item We study the effect of the patch size on a crack segmentation performance.

\end{itemize}
\subsection{Structure of the Article}
In Section \ref{sec:Method} we describe the proposed deep learning method for crack segmentation. 
Section \ref{sec:Datasets} of this paper is dedicated to the datasets on which we conduct our experiments including the novel \textit{Cracks in Steel Bridges} (CSB) dataset. 
Section \ref{sec:Experiments} explains the used training algorithms and evaluation approaches. 
In Section \ref{sec:Results and discussion} we summarise the experiments and discuss the results obtained.


\section{Models and Methods}
\label{sec:Method}
The proposed algorithm uses an encoder-decoder network to segment images of cracks in steel bridges.  
The encoder-decoder network takes an image as input and produces a segmentation map as an output, which assigns a value of 0 (background) or 1 (crack) to each pixel of the input image\footnote{Throughout the paper, segmentation maps are visualized with inverted colors - white background and a black crack.}.  However, an entire image cannot be processed by the encoder-decoder network at once, due to hardware computing memory constraints and the fact that the images in our dataset have dimensions of up to 4608x3456 pixels. We estimate that training of our neural network on images of this size would require 189 GB of RAM, while the GPUs we used for experiments are NVIDIA A100 with RAM of 40 GB.
One possible solution to this problem would be to downsample the input images to a smaller size in order to allow efficient segmentation. 
However, downsampling of an image can lead to a loss of fine details including cracks since they are thin structures of the images, with a width of a few pixels, depending on the spatial resolution of the image, i.e. pixel to millimeter ratio. 
Therefore, the downsampling of the images could downgrade segmentation performance significantly.
A common solution to this problem of working with images of very large resolution and size is to decompose them into patches that can be processed efficiently.
In this paper, we choose the patch size to be 512x512 pixels. 
Each patch of the image is then segmented separately and the resulting segmentation maps are stitched together to produce a global segmentation of the whole image.  

CNN's \cite{lecun1998CNN,krizhevsky2017imagenet} are powerful image processing tools developed and widely used for image classification, and segmentation, amongst other applications. It is common to distinguish parts of a CNN into a feature encoder, and classifier/ output layer.
The CNN encoder involves multiple layers of filters and down-samplers to hierachically generate feature maps, that encode important semantic image information.
The filters of the encoder extract geometrical features of an image, whilst the downsamplers decrease the spatial dimension of the feature maps, thus, providing a bigger receptive field \cite{luo2016understandingERF} of filters deeper in the neural network and reducing computational load. 

To perform pixel-to-pixel image segmentation fully convolutional networks (FCN) were proposed \cite{long2015FCN} as an extension of CNNs. 
In the original FCN \cite{long2015FCN}, the classifier of the CNN is replaced with an up-sampling layer transforming the encoded feature maps with low spatial dimension and rich semantic information into a segmentation map.

U-Net architecture \cite{ronneberger2015unet} extends the original FCN by replacing its relatively simple upsampling layer with a multi-layer decoder. 
A more complex decoder of the U-Net, compared to the one in the original FCN, allows gradual spatial upsampling of the feature maps and more efficient conversion of the feature maps into an output segmentation map. 
The long-range skip connections between the encoder and decoder introduced in \cite{ronneberger2015unet} help to restore spatial information that would be lost due to downsampling in the encoder.

A recent example of adopting the encoder-decoder architecture for the cracks segmentation task was proposed by Konig et al. \cite{Konig2021OED}. 
This method shows state-of-the-art performance on the CFD dataset \cite{shi2016CFD} 
according to Table \ref{tab:CFD other works results: 2pixel} in the \ref{app:B}. 
Therefore, we will use this method as our primary baseline over which we develop our architectural and training strategies leading to significant improvements.

\begin{figure*}[t]
    \centering
    \includegraphics[width=1\linewidth]{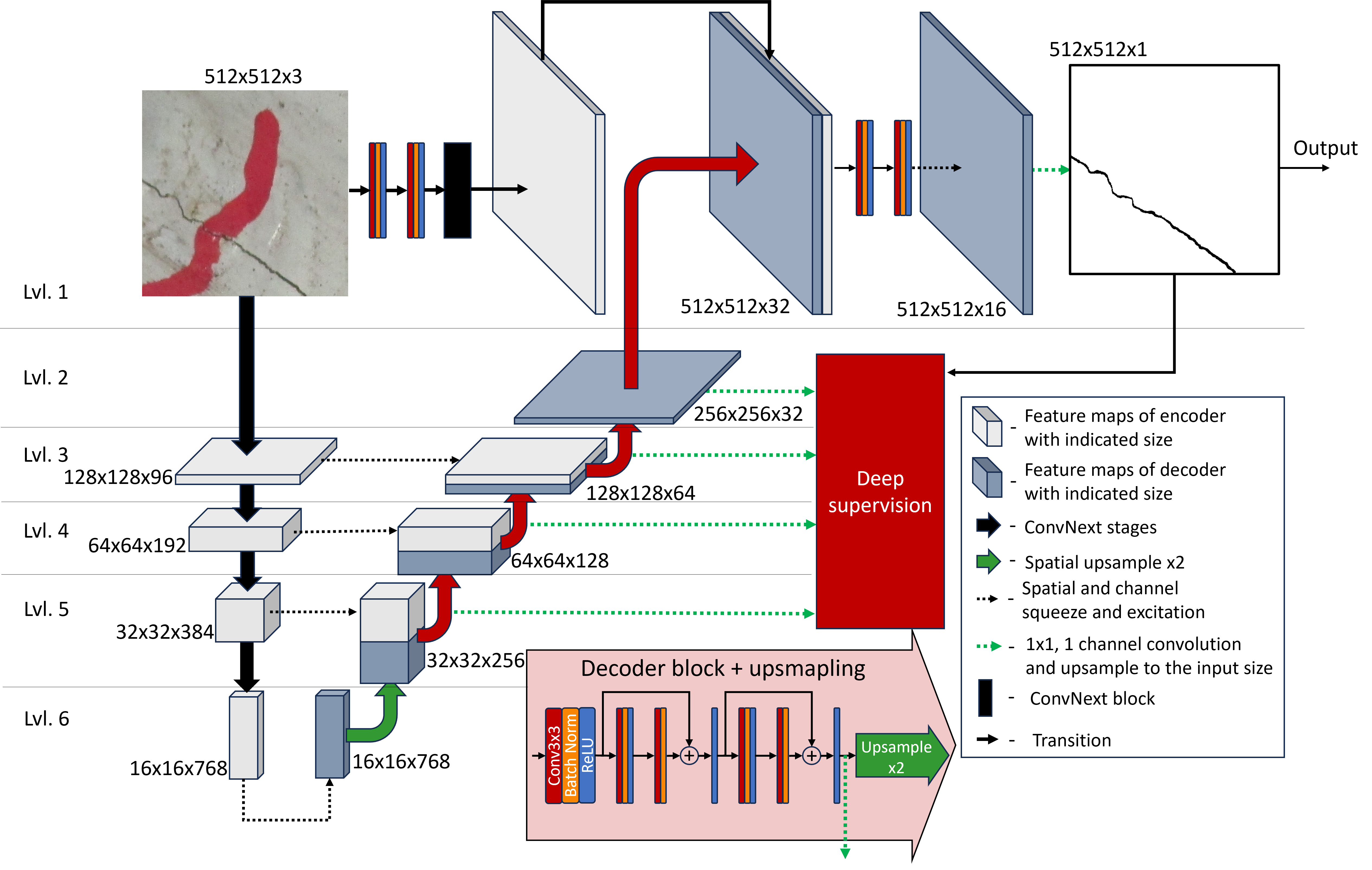}
    \caption{Scheme of the proposed encoder-decoder network}
    \label{fig:Network scheme}
\end{figure*}

First,  we propose the use of the recent ConvNext \cite{Liu_2022ConvNext} architecture as the network encoder. 
Secondly, we further enhance the architecture by implementing spatial, channel squeeze, and excitation layers \cite{roy2018scSE}. Finally, we modify a few layers in the last (i.e. high scale) layers by making skip connections in order to enable better processing of the fine-grained details in the images. Figure \ref{fig:Network scheme} depicts the schematic for the overall encoder-decoder network architecture. 

\subsection{Encoder}
Several popular pre-trained CNNs such as VGG \cite{simonyan2014VGG}, Inception \cite{szegedy2016Inception}, MobileNet \cite{howard2019MobileNetV3}, and ResNet \cite{he2016resnets} are made publicly available. 
The weights of these pre-trained neural networks are optimized on a large classification dataset such as ImageNet \cite{russakovsky2015imagenet}, which contains about 1.2 million images categorized in 1000 classes. It is a common approach to use the pre-trained feature extractors of these neural networks as encoders in the encoder-decoder networks. This allows having the general learned features of these pre-trained neural networks, which can be applied to different downstream tasks with a little fine-tuning. This approach is called `transfer learning' \cite{yosinski2014transferlearning}. 

Despite the fact that the ImageNet dataset consists of generic objects, the features learned on the ImageNet can be successfully applied to crack detection with fine-tuning as shown in \cite{bukhsh2021TransferLearningCrackDetection} where cross-domain transfer learning is applied.    

In \cite{Konig2021OED} experiments were done, exploring the effect of different popular CNNs, namely VGG 19 \cite{simonyan2014VGG}, EfficientNets \cite{tan2019EfficientnetNet} and ResNet 50 \cite{he2016resnets} on our baseline neural network performance. In 2022 a new powerful ConvNext \cite{Liu_2022ConvNext} was proposed, which showed state-of-the-art performance on the ImageNet \cite{russakovsky2015imagenet} dataset. It was a natural decision to use ConvNext as an encoder in the encoder-decoder network for crack segmentation. This is what we did in our work, using the 'Tiny' version of the ConvNext. 

The ConvNext modifies ResNet by using numerous design improvements that were already shown to be useful separately in other works. 
These design improvements concern different levels of design, ranging from macro design, for example, stage ratio, to micro design, such as change of activation functions (for details, please refer to \cite{Liu_2022ConvNext}). 
Moreover, improved optimization algorithms were used for the ConvNext, which we also use in our work. 
After the publication of the ConvNext in 2022, in 2023 a ConvNext V2 was released \cite{woo2023ConvNextv2}, which showed even better results by using self-supervised pretraining. 
However, in this work, we conduct our experiments only with ConvNext, and the reader should keep in mind, that replacing the ConvNext with ConvNext V2 has the potential to produce even better results for crack segmentation. 

The ConvNext consists of 4 stages, which are represented by the thick black arrows in Figure \ref{fig:Network scheme}. 
In our neural network the the ConvNext is used to take as input an image patch and to produce its feature maps, which are then used as input to the decoder.

\subsection{Decoder}
The architecture of our decoder is based on the one proposed in \cite{Konig2021OED}. 
The scheme of our decoder can be seen in Figure \ref{fig:Network scheme}. It is divided into 6 levels. 
At level 6  the feature maps generated by the encoder are first processed by the channel and spatial squeeze and excitation module \cite{roy2018scSE}. 
This module is often also referred to as an attention module because it assigns learned weights to particular locations and channels of feature maps, thus emphasizing them for the subsequent parts of the neural network, in other words - it helps to give attention to the most relevant information.
Afterwards, the weighed feature maps are spatially upsampled by a factor of two using the nearest neighborhood method. 

The feature maps produced by the decoder at level 6 are passed to level 5 of the decoder.
At level 5, these feature maps are concatenated with feature maps of level 5 of the encoder, to which the channel and spatial squeeze and excitation module is applied. The concatenated feature maps are processed by a convolution layer with a kernel size 3x3 and a stride of 1 to decrease the number of feature maps channels, which is increased after concatenation, to the number of channels of the decoder feature maps at level 5. The dimensions of the feature maps of the decoder as well as of the encoder are indicated in Figure \ref{fig:Network scheme}. 
The convolution is followed by batch normalization and the ReLU activation function. 
After the channel decrease operations, a residual block is added as shown in Figure \ref{fig:Network scheme}. 
The residual block that we use consists of convolution layers with kernel size 3x3, batch normalization layers, and ReLU activation functions, with two short-range skip connections. 
The detailed scheme of the decoder block is shown in Figure \ref{fig:Network scheme}. 

Decoder operations at levels 4-2 are identical to the decoder operation at level 5.
The decoder block at level 2 is identical to the previous one with the only difference that there is no concatenation with feature maps from the encoder. 
The reason for this is that the first layer of the ConvNext downsamples the input image by a factor of four, thus skipping level 2 of our neural network.
After the upsampling at level 2, feature maps reached the spatial size of the input image at level 1 of the decoder. 
These feature maps are concatenated with feature maps generated from the input image without any downsampling. This is done to restore information with the highest frequencies which could be lost due to the downsampling operations in the encoder. To produce feature maps from the image at Level 1, 2 convolutions are applied with kernel size 3x3, each followed by batch normalization and ReLU activation function. Afterwards, a single ConvNext block \cite{Liu_2022ConvNext} is applied with 16 channels. 
Finally, the sequence of 3x3 convolution, batch normalization, and ReLU activation functions is applied twice. 
The feature maps generated by the decoder are sent through the pointwise convolution followed by the sigmoid activation function to generate the final segmentation map.

\subsection{Loss function with inversion}
The loss function provides an objective for the optimization of neural network parameters. A properly formulated loss function should give an accurate estimate of errors, which a neural network makes on an entire range of training and testing examples.  

The summation of the BCE loss and the Dice loss was proven to be a good loss function for crack segmentation \cite{yang2021automatic} and is, indeed, a good candidate to be used in our case. However, as will be shown below, this loss function does not give an accurate error estimate, when applied to image patches without cracks, where each pixel in the ground truth annotation is equal to 0. Since in our work, we use a substantial amount of image patches without cracks, we present a concept of loss function inversion, which allows to solve the aforementioned problem and gives good learning objectives for neural network training on both, crack patches and background patches.

\subsubsection{Sum of Dice loss and BCE loss}
Since cracks are thin compared to the size of the image, the number of crack pixels is much lower than the number of background pixels in the annotated segmentation maps. 
The Dice loss function \cite{sudre2017DICE_loss} was introduced to deal with this data imbalance problem. Furthermore, in \cite{yang2021automatic} adding Binary Cross Entropy (BCE) loss to the Dice loss allowed authors to get performance gain. Interestingly in \cite{yang2021automatic} it was also found that Dice loss alone gives very "sharp" segmentation maps, where each value of the output segmentation map is either close to 0 or to 1. But the sum of the Dice loss and BCE loss gives smoother segmentation maps, with more intermediate values. This, gives more flexibility in the choice of the threshold value, allowing to balance the false positives and false negatives of the crack detection algorithm. The Dice loss and the BCE loss can be calculated as follows:
\begin{equation}\label{eq:BCE}
\mathcal{L}_{BCE}(\mathbf{y},\hat{\mathbf{y}}) = -\frac{1}{N}\sum\limits_{i=1}^N  \Bigl[ {y_i \log{\hat{y_i}} + (1-y_i)(\log(1-\hat{y_i}))} \Bigr],
\end{equation}

\begin{equation}\label{eq:Dice}
\mathcal{L}_{Dice}(\mathbf{y},\hat{\mathbf{y}}) = 1 - \frac{\sum \limits_{i=1}^N{y_i\hat{y_i}}+\epsilon}{\sum \limits_{i=1}^N{y_i}+\sum \limits_{i=1}^N{\hat{y_i}} + \epsilon},
\end{equation}
where N is the number of pixels in the image, and $y_i$ and $\hat{y_i}$ are values of the $i$-th pixel in the ground-truth and predicted segmentation maps respectively, that we store in the input vectors $\mathbf{y}=(y_i)_{i=1}^N$ and $\hat{\mathbf{y}}=(\hat{y}_i)_{i=1}^N$, with $y_i \in [0,1]$ and $\hat{y}_i \in \{0,1\}$.  The $\epsilon$ is added to avoid division by zero and is set to 1. The sum of the BCE loss and Dice loss is often used as a loss function for a crack segmentor:
\begin{equation} \label{eq:Dice + BCE}
\mathcal{L}(\mathbf{y},\hat{\mathbf{y}})= \mathcal{L}_{BCE}(\mathbf{y},\hat{\mathbf{y}}) + \mathcal{L}_{Dice}(\mathbf{y},\hat{\mathbf{y}}).
\end{equation}

\subsubsection{Loss function with inversion}
\label{sec:Methods:loss function with inversion}
As will be explained in Section \ref{sec:Datasets}, during the training of the neural network on the images of cracks in steel bridges we use lots of background patches, all pixels of which are annotated as background, i.e. $\mathbf{y}=\mathbf{0} \leftrightarrow \sum_{i=1}^N{y_i}=0$, since $y_i\in [0,1]$. 
For a background patch, the Dice loss function as presented in  Eq.~(\ref{eq:Dice}) then becomes:
\begin{equation}\label{eq:Dice background}
\sum \limits_{i=1}^N{y_i}=0
\Rightarrow
\mathcal{L}_{Dice}(\mathbf{y},\hat{\mathbf{y}}) = 1 - \frac{0+\epsilon}{\sum \limits_{i=1}^N{\hat{y_i}}+0 + \epsilon}
\end{equation}
Thus, $\mathcal{L}_{Dice} \simeq 1$ irrespective of the predicted segmentation map $ \hat{\mathbf{y}}$, 
whenever $\sum \limits_{i=1}^N{\hat{y_i}} \gg \epsilon$. 
A visible blob on an image takes a significant number of pixels and therefore, in the case of the noticeable false positive segmentation the inequality   $\sum \limits_{i=1}^N{\hat{y_i}} \gg \epsilon$ holds.
This means that for a background patch, different predictions of the neural network have almost no effect on the value of the Dice loss.
The only exception is when the neural network marks all pixels as background pixels, then the Dice loss naturally equals 0. 

To tackle the problem described above and to make the Dice loss sensitive to neural network errors on the background patches, we introduce a loss function with the inclusion of inversion for background patches:

\begin{equation}
    \mathcal{L}_{inversion}(\mathbf{y},\hat{\mathbf{y}})= \begin{cases}
    \mathcal{L}(\;
   \textrm{inv}(\mathbf{y})\,,\,\textrm{inv}(\hat{\mathbf{y}})\;),& \text{if } \sum \limits_{i=1}^N{y_i}=0\\
    \mathcal{L}(\mathbf{y},\hat{\mathbf{y}}),              & \text{otherwise}
\end{cases}
\end{equation}
where label inversions are given by $\textrm{inv}(y)=-y+1$, so that $\textrm{inv}(0)=1$ and $\textrm{inv}(1)=0$, stored in 
$\textrm{inv}(\mathbf{y})=
    (\textrm{inv}(y_i))_{i=1}^N
$,
and
where $ \mathcal{L}$ is the loss function defined by Eq.~(\ref{eq:Dice + BCE}).
It can be interpreted as follows. When during training of the neural network a crack patch is encountered, we apply the normal loss function as given by Eq.~(\ref{eq:Dice + BCE}) which estimates the loss for crack segmentation. But, when a background patch is encountered ($\mathbf{y}=\mathbf{0}$), we invert 0-s and 1-s in the ground truth segmentation map and in the predicted segmentation map. In this inverted segmentation map, the background pixels are marked with 1, and the crack pixels are marked with 0. Thus the inverted loss function asses errors of a background segmentor rather than a crack segmentor as in the crack patch case. Thereby we are able to avoid the 0+$\epsilon$ in the numerator of the Dice loss (Eq.~(\ref{eq:Dice})) and the mistakes in segmentation made by the neural network have a sensible effect on the loss function value, thus, giving an accurate estimate of errors on the background patches.  

The introduced loss function with inversion allowed us to better train the proposed neural network using a large number of background patches. This helped to significantly decrease the false positive rate, which otherwise have a negative effect on the overall performance of the neural network for crack segmentation.

\subsection{Deep supervision}
Deep supervision \cite{lee2015deepsupervision} is employed in our work to improve the segmentation results. 
Deep supervision helps to avoid exploding/vanishing gradients problems and to learn more robust and discriminative features in the intermediate layers of the deep network, thus, accelerating neural network training and improving its performance.
To follow the deep supervision approach, apart from the segmentation map produced by the last stage of the decoder (target segmentation map), additional segmentation maps are generated at levels 2-5 of the decoder (companion segmentation maps). 
To produce companion segmentation maps, each of the feature maps resulting from levels 2-5 after the last activation function in the decoder
block is sent through point-wise convolutions with a single channel. 
Then it is upsampled to the size of the input image using the nearest neighborhood method and processed by the sigmoid activation function. 
These operations are shown with a green dotted line in Figure \ref{fig:Network scheme}.
For each of the companion segmentation maps loss is computed in the same way as for the target segmentation map.
Thus, when deep supervision is applied the total loss function becomes:
\begin{equation}\label{eq:Deep Supervision}
\mathcal{L}^{total}(\mathbf{y},\hat{\mathbf{y}}) = \mathcal{L}^{target}(\mathbf{y},\hat{\mathbf{y}}) + \sum_{i=0}^{n} w_i \mathcal{L}^{companion}_{i}(\mathbf{y},\hat{\mathbf{y}_{i}}) ,
\end{equation}
where $n$ is the number of companion segmentation maps, $w_i $ is the weight of each companion loss and $\hat{\mathbf{y}_{i}}$ is a vector representing $i$-th companion output segmentation map. We set each weight to 1.

\section{Datasets}
\label{sec:Datasets}
Through this paper we use two datasets: an open-source CrackForest dataset (CFD) \cite{shi2016CFD} and our own Cracks in Steel Bridges (CSB) dataset which is firstly presented to the public in this work and described in this section of the paper.

The CFD dataset was extensively used previously for the development of deep learning methods for crack segmentation, therefore, experiments on the CFD dataset are carried out to give a fair comparison of the proposed algorithm with earlier methods for crack segmentation.

Furthermore, we present the Cracks in Steel Bridges (CSB) dataset. The proposed deep learning algorithm and earlier algorithms are trained and tested on this novel and challenging dataset that can be downloaded from  \cite{CSB_dataset}.

\subsection{CFD}
\begin{figure}
    \centering
    \includegraphics[width=1\linewidth]{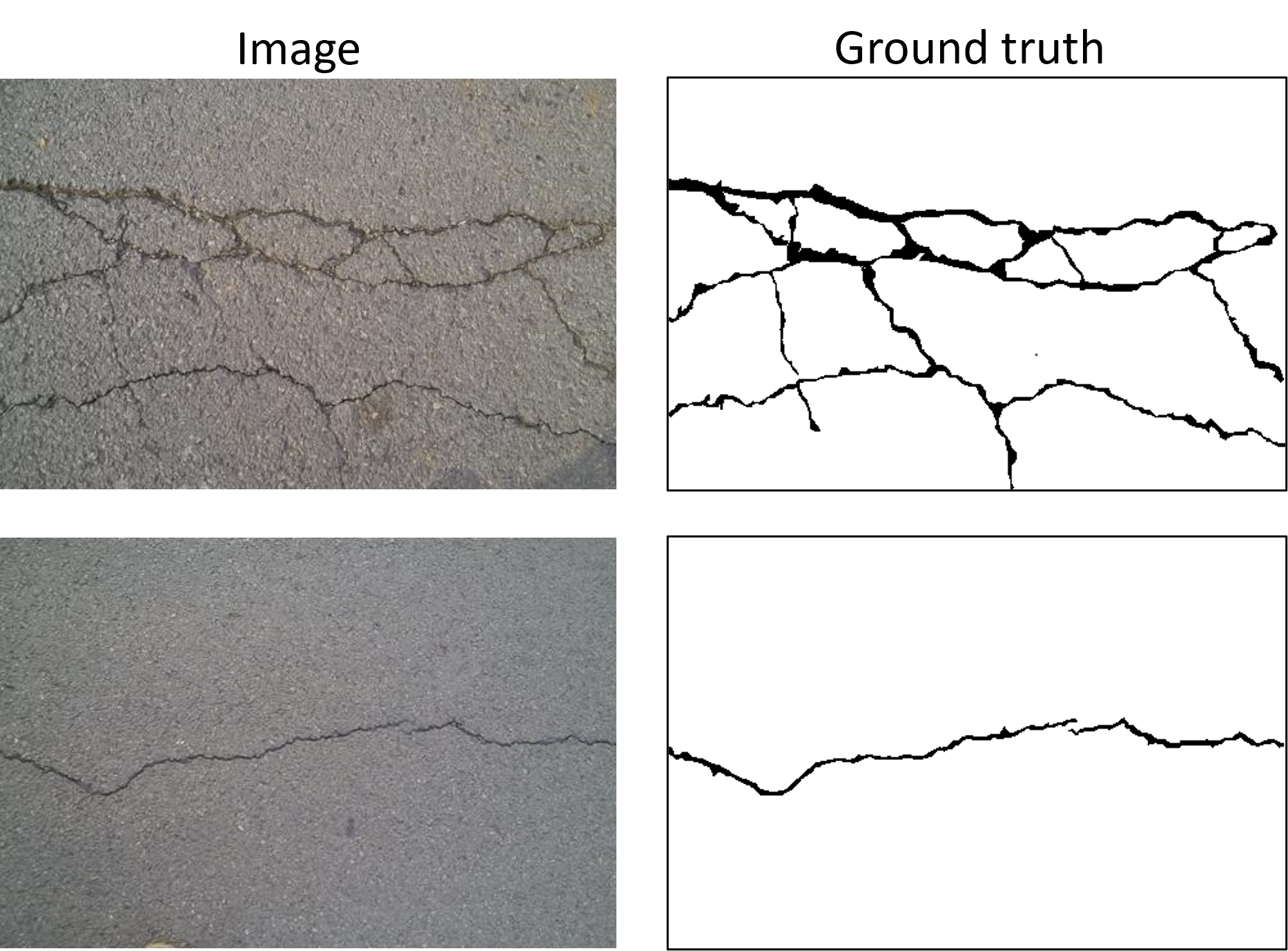}
    \caption{Examples of images and ground truth crack segmentation from the CFD dataset}
    \label{fig:CFD examples}
\end{figure}

CrackForest Dataset (CFD) is an open-source dataset that was introduced in \cite{shi2016CFD}. It contains 118 images of size 480x320 pixels with cracks in the road surfaces. 
The images were collected in Beijing using a smartphone camera. 
For each image, a ground truth segmentation map was created by means of manual annotation. 
Apart from cracks, the images capture shadows, oil spots, and water stains. 
In \cite{kang2021efficient} the complexity of the CFD and the complexity of 5 analogous datasets were quantitatively compared. 
A number of separate objects recognized by Felzenszwalb’s graph segmentation algorithm \cite{Felzenszwalb2004EfficientGI} was used as a basis for quantitative complexity estimate. 
According to this comparison, the CFD is the simplest dataset. 
However, despite its low estimated complexity it can be challenging to develop an algorithm that gives accurate results, because of differences in the crack sizes and inconsistency in the annotation. Thin dark lines with similar appearances in some images are annotated as cracks and in some images as background. 
Examples of the images from the CFD dataset can be seen in Figure \ref{fig:CFD examples}. 

Following a common practice \cite{Konig2021OED,Wj2019FPCNet,shi2016CFD} the images from the CFD dataset were split into train and test sets containing 60\% and 40\% respectively. 
Thanks to the fact that images in the dataset are enumerated, similar to \cite{Konig2021OED}, in our work, the images from 1 to 72 are used as train images and the rest as test images. 
As was pointed out in \cite{Konig2021OED}, image number 42 has a wrong ground truth annotation and was not used for the experiments. 

\subsection{Cracks in the Steel Bridges (CSB) dataset}

The CSB dataset consists of 755 images with cracks and 300 images without cracks which were collected during regular bridge inspections. The images have various sizes with a maximum of 4608x3456 pixels.
They were captured at varying distances and angles relative to the bridge, ranging from less than 0.5 meters up to 5 meters, as estimated based on a visual assessment. 
Apart from cracks, corrosion also can be seen and in some images, corrosion is developed around a crack and obscures it to some extent. Often, in the corroded spots paint is intentionally removed around cracks for inspection purposes.  On many images, physical inspector markings are presented, indicating crack tips, their size, and inspection dates. Moreover, since these are images of structures in use, dust, dirt, raindrop spills, bugs, and spider webs are present. Many images were taken using artificial light sources to improve visibility. The surfaces of the bridges in the images are covered with paint of different colors.

Pixel-wise annotations of the images are made using a semi-automatic tool specifically developed for the purpose \cite{CP1}, which relies on a geometric tracking algorithm \cite{duits2018optimalpath}, described in \cite{CP1} and its implementation is available in \cite{github_segmentation_tool}. A short summary of the algorithm is given in \ref{app:A} of this paper. 
The tool uses manual input indicating the position of two crack endpoints on an image. 
The geometric tracking algorithm uses this input as a start and end point to find a path of the crack between them. 
Based on this track, the segmentation of the crack is done automatically.

Making use of this algorithm allowed reducing annotation time from 30 minutes to less than 1 minute per image.
However, this time savings come with the cost of a reduction in annotation accuracy. In \cite{CP1} it was shown that the tool 
performs at the level of 83 \% accuracy of manual annotation measured by means of $F1$-score, see \ref{sec:Experiments:Evaluation}. 
A sophisticated study of the effect of annotation errors on a neural network for crack segmentation was done in \cite{xu2023label_errors}. 
It was shown that 20\% of annotation errors give up to a 6\% drop in neural network performance measured by means of $F1$-score. In Section \ref{sec:Results and discussion} of this paper we investigate the effect of annotation errors on neural network performance, made by the semi-automatic annotation tool, which was used to annotate the CSB dataset. We found that these errors cause a drop in $F1$-score of no more than 2\%.

\begin{figure}
    \centering
    \includegraphics[width=1\linewidth]{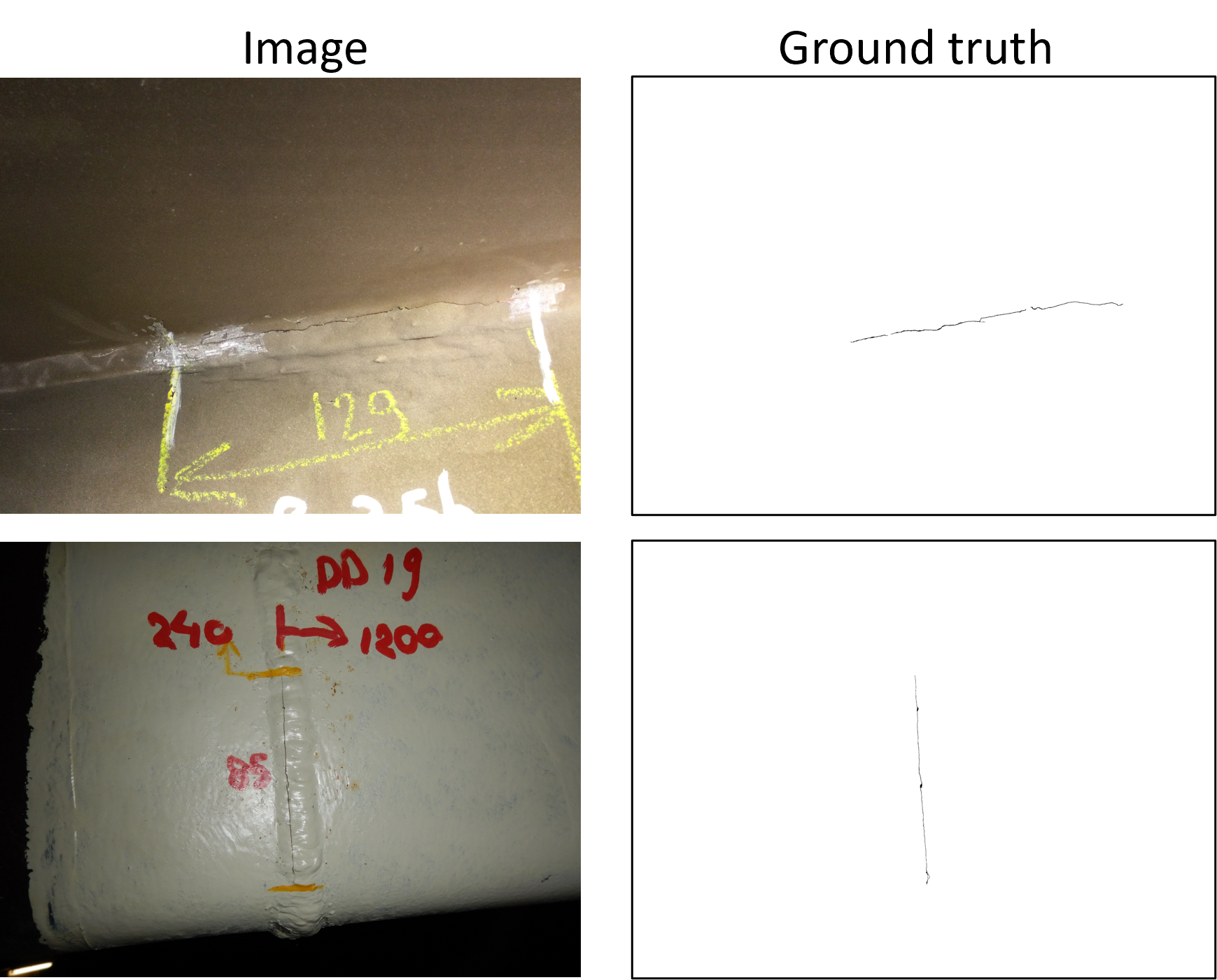}
    \caption{Examples of images and ground truth crack segmentation from the CSB dataset}
    \label{fig:CSB examples}
\end{figure}

\begin{figure*}
    \centering
    \includegraphics[width=0.9\linewidth]{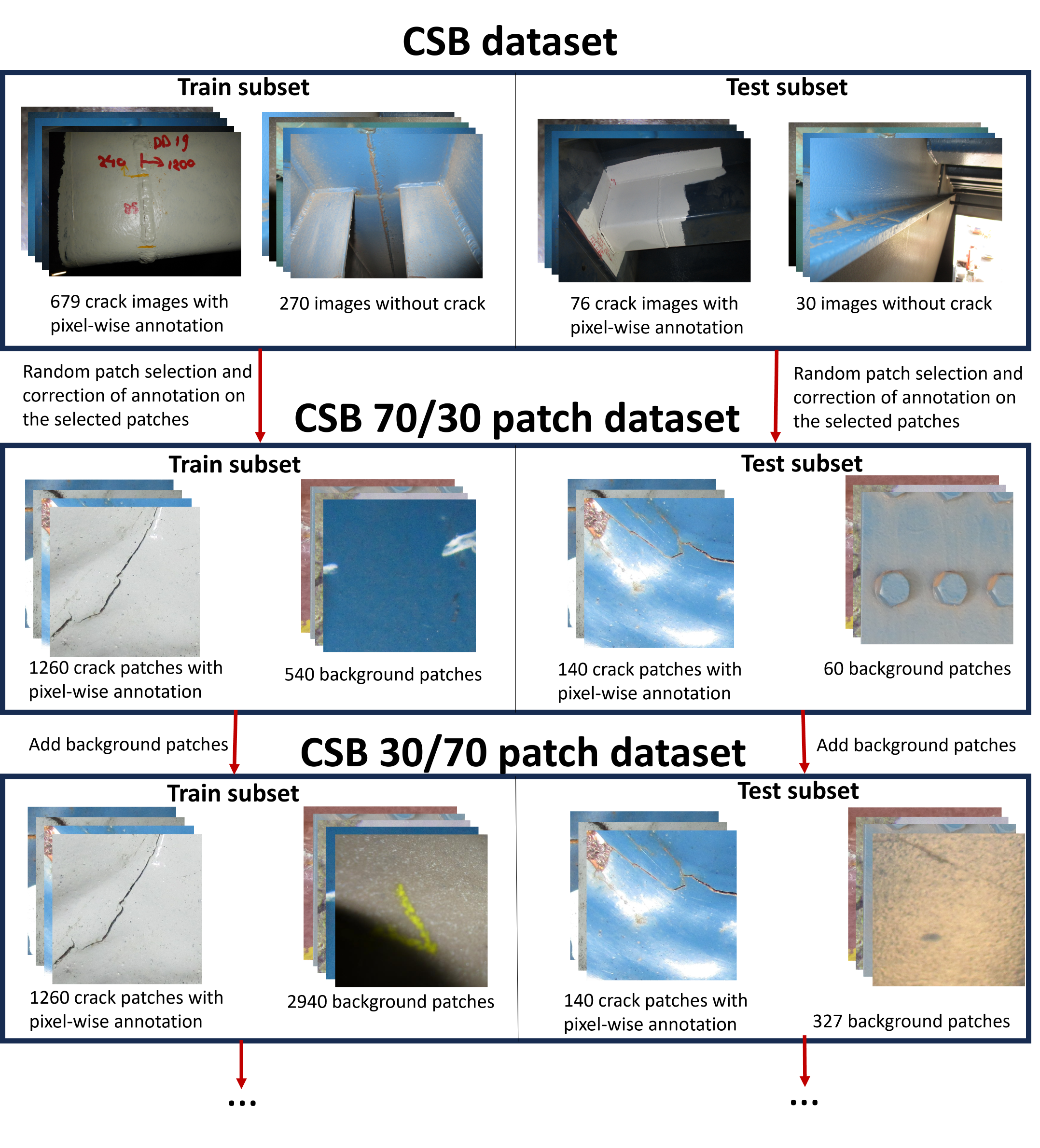}
    \caption{Illustration of the datasets structure}
    \label{fig:Patch dataset}
\end{figure*}

Examples of the images from the CSB dataset with ground truth pixel-wise annotation are shown in Figure \ref{fig:CSB examples}. Together with the 755 annotated images with cracks, we use additional 300 images of bridges without any cracks. All 1055 images were split into test and train subsets containing 10\% and 90\% of images respectively. The structure of the CSB dataset is visualized in Figure \ref{fig:Patch dataset}.

\subsection{CSB patch datasets}

\begin{figure}
    \centering
    \includegraphics[width=1\linewidth]{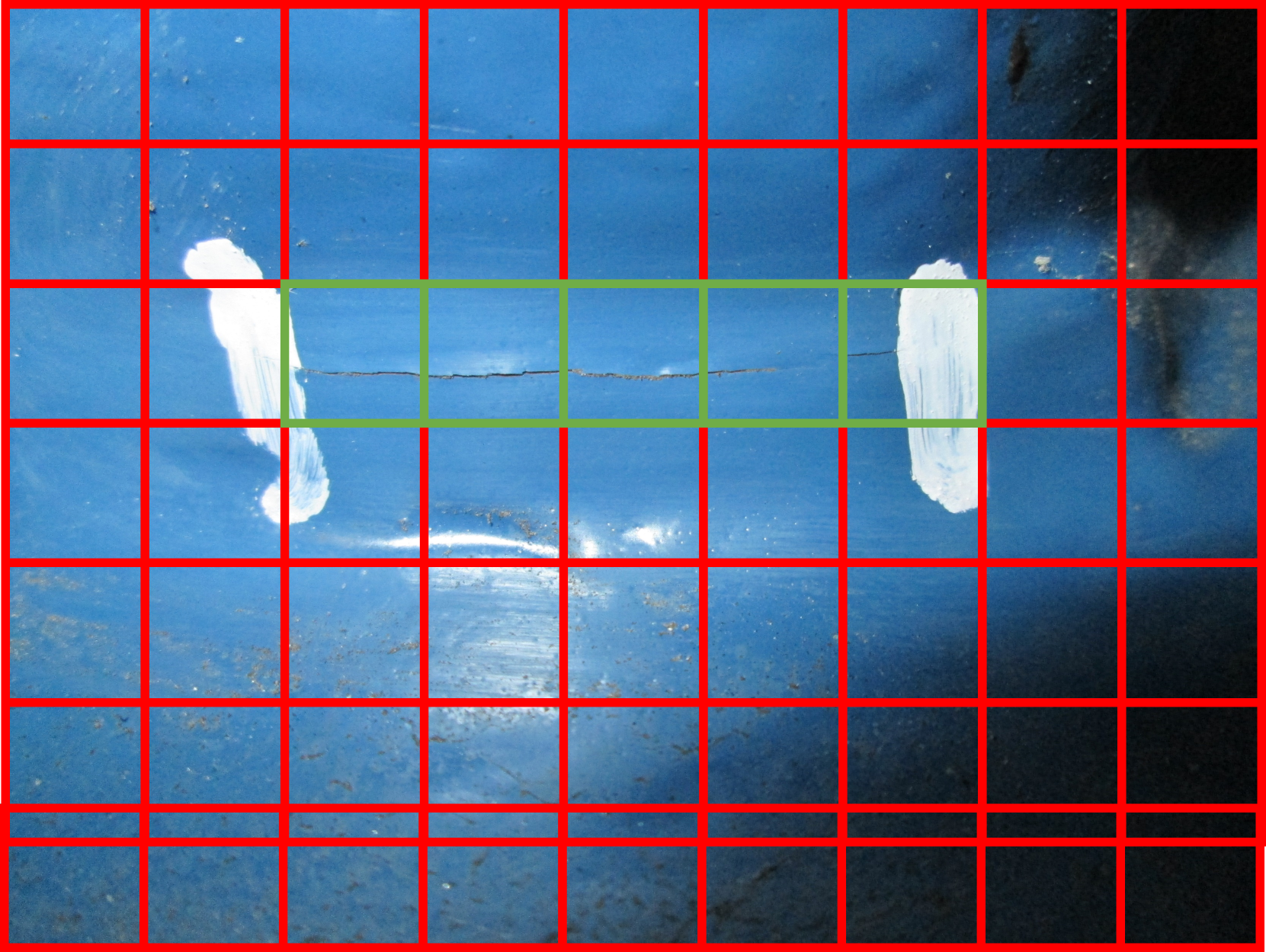}
    \caption{Example of an image of size 4608x3456 pixels split into patches of size 512x512. Red squares represent background patches and green squares represent crack patches. Note: the bottom row of patches overlaps with the row above it}
    \label{fig:Patch full split}
\end{figure}

The size of the images from the CSB dataset is too large to be processed by an encoder-decoder network at once, therefore the images have to be split into patches. As shown in Figure \ref{fig:Patch full split}, all the images are divided into non-overlapping patches (only the last rows and/or columns of the patches may overlap) that are used to train and
test the neural network presented in Section \ref{sec:Method}. However, the area of the cracks is small compared to the area of entire images, and when all images from the CSB dataset are split into patches, only about 6.45\% (3 950) of these patches contain cracks while the rest 93.55\% (57 276) contains only background. This patch imbalance problem can be seen in Figure \ref{fig:Patch full split}, where an entire image of size 4608x3456 pixels is split into patches of size 512x512 and only 5 of 63 patches, which is approximately 8\%, contain the crack.

Training of a neural network on the patch dataset having this crack-to-background patches ratio causes the gradient descent algorithm to stacking at a local minimum in the optimization space, where the neural network learns to ignore cracks in any image. We call this an "all-white segmentation maps" problem.
During our initial experiments with a simple U-Net architecture, we found that 70\% to 30\% of crack-to-background patches in the train set allow to avoid this issue. 

Therefore, from the CSB dataset, we generate the CSB 70/30 patch dataset. It constitutes of 1400 random crack patches and 600 random background patches taken from the images of the CSB datasets. Patches taken from train images of the CSB dataset are assigned to the train subset of the CSB 70/30 patch dataset and patches taken from test images of the CSB dataset are assigned to the test subset of the CSB 70/30 patch dataset. Moreover, the ratio of the number of crack patches to background patches in both the test and train subsets of the CSB 70/30 patch dataset is also 70\%/30\%.
The structure of the CSB 70/30 patch dataset and its relation to the CSB dataset is illustrated in Figure \ref{fig:Patch dataset}. To provide pixel-wise crack annotation, the semi-automatic crack segmentation tool mentioned before \cite{CP1} was used. As the tool provides segmentation with errors, the annotations for the  1260 crack train patches and 140 crack test patches were manually corrected. 

Furthermore, we conduct experiments to investigate the effect of the crack patches to background patches ratio used for training on the performance of the neural network, when tested on the entire CSB dataset. To do so, we generate a few more datasets with different crack-to-background patch ratios.

To generate the CSB 30/70 dataset we take the CSB 70/30 patch dataset and add to them new random background patches, increasing the fraction of the background patches to the 70\%. This idea is depicted in Figure \ref{fig:Patch dataset}. 
To generate the CSB 10/90 patch dataset we add random background patches to the CSB 30/70 patch dataset.

The idea of adding patch datasets resonates with the practical application, where new images of cracks in steel bridges are not always available and it is much easier to get additional images of a bridge structure without a crack.
To summarise, the following datasets of steel bridge images were used in this study: 
\begin{itemize}
\item \textbf{CSB dataset}: all patches from entire images are used, giving 6.45\% (3950) crack patches and 93.55\% (57276) background patches;
\item \textbf{CSB 70/30 patch dataset}: 1400 (70\%) crack and 600 (30\%) background patches;
\item \textbf{CSB 30/70 patch dataset} 1400 (30\%) crack patches and 3267 (70\%) background patches;
\item \textbf{CSB 10/90 patch dataset} 1400 (10\%) crack patches and 12600 (90\%) background patches.
\end{itemize}


\section{Training and evaluation methods}
\label{sec:Experiments}

\subsection{Training}
For the training of all neural networks, identical training settings are used. 
For the optimization of the neural network weights an "ADAMW"  \cite{loshchilov2017adamw} algorithm is used. The initial learning rate is set to $ \lambda_{initial} = 0.001$ with an exponential learning rate decay schedule: 
\begin{equation}
\lambda = \lambda_{initial} \cdot \gamma^{e}, 
\end{equation}
where $\lambda$ is the learning rate, $\gamma = 0.99$ and $ e$ is current epoch number. Furthermore, to make better use of the pre-trained weights of the ConvNext, we apply the stage-wise learning rate decay technique:
\begin{equation}
\lambda_{stage}=\lambda \cdot k^{N+1-n},
\end{equation}
where $k=0.7$ is the selected parameter, $N=4$ is the total number of the ConvNext stages and n is the stage number, for which the learning rate is computed.  The coefficient $k$ was chosen empirically, such that it gives good performance on the CSB dataset. 

The weight regularisation parameter for the "ADAMW" optimization algorithm is set to $1\cdot10^{-5}$. 

We apply data augmentation techniques artificially increasing the number of images in the dataset. Each time a patch was picked from the dataset to be used for the neural network training, the following augmentation steps were applied to the patch:
\begin{itemize}
\item Horizontal flip with 50\% probability and vertical flip with 50\% probability;
\item Random rotation by 0, 90, 180, or 270 degrees;
\item Random crop to get the image patch (used only for CFD datasets to get 288x288 pixels patch);
\item Random adjustment of image brightness, contrast, and saturation in the range from 75\% to 125\%;
\item Random image hue adjustment in the range -10\% to +10\%;
\item With a 50\% probability addition of random image noise, by multiplying each pixel by a value randomly sampled in the range between 0.9 and 1.1;
\item With a probability of 40\% one of the following:

\begin{itemize}
\item random spatial reduction of the image with a scale factor in the range from 75\% to 125\%, with subsequent padding with mirroring to keep the patch dimension constant;
\item zoom in on the image with a scale in the range from 75\% to 100\%, with subsequent interpolation, to keep the patch size the same.
\end{itemize}

\end{itemize}

\subsection{Evaluation}
\label{sec:Experiments:Evaluation}
The re-implemented and the proposed algorithms were evaluated on the test data subsets.
It is common practice to evaluate the performance of the crack segmentation algorithm by means of precision ($Pr$), recall ($Re$), and $F1$-score ($F1$). To calculate these metrics, the number of True Positive ($TP$), False Positive ($FP$), and False Negative ($FN$) pixels in the segmentation map is required:
\begin{equation} \label{eq:Precision}
Pr = \frac{TP}{TP+FP},
\end{equation}
\begin{equation} \label{eq:Recall}
Re = \frac{TP}{TP+FN},
\end{equation}
    \begin{equation} \label{eq:F1-score}
F1 = \frac{2 \cdot Pr \cdot Re}{Pr + Re} = \frac{2 \cdot TP}{2 \cdot TP + FP + FN}.
\end{equation}

Precision may be interpreted as a fraction of correctly segmented crack pixels among all pixels segmented as a crack by an algorithm. Recall may be interpreted as correctly segmented crack pixels among all crack pixels according to a ground truth annotation. Finally, the $F1$-score is a harmonic mean of precision and recall. 
Often, precision, recall, and $F1$-score are calculated separately for each image in a test set and mean values are taken to show the overall performance. However, this assessment approach may not give an accurate performance estimate for datasets, where the amount of crack pixels varies significantly from image to image, and/or where the number of background patches is significantly larger than the number of crack patches as in the CSB dataset. 
In this case, an image without crack pixels in the ground truth annotation results in $TP$=0 and $FN$=0, meaning that $F1$=0 is insensitive to the number of false positives. This is the same problem as described in Section \ref{sec:Methods:loss function with inversion}, because the $F1$-score is equivalent to the Dice score for two class segmentation.  
To cope with this problem, in our evaluation, we sum the number of $TP$, $FP$, $FN$ for the individual images in the test set, and based on them we calculate an aggregate precision, recall, and $F1$-score for the entire test set. 

All the neural network architecture considered in the context of this article produces a segmentation map with values between 0 and 1, but counting the number of $TP$, $FP$, and $FN$  requires binary segmentation maps. To produce binary segmentation maps on the test images, a threshold value should be chosen.
In our work, the threshold value is always set to 0.5.

\subsubsection{Evaluation on CFD}
\label{subsec:Evaluation on CFD}
In \cite{Konig2021OED,shi2016CFD,Wj2019FPCNet} a less rigorous evaluation of the performance of the neural network architecture is made where both $FP$ and $FN$ pixels lying within a tolerance of 2 pixels from the ground truth annotation are counted as $TP$. This is in contrast with common practice in which a tolerance region around the ground truth does not exist. This is done to account for the fact that often crack edges are blurred and it may be difficult to provide a single correct binary segmentation map. To compare the implemented algorithms with the results presented in earlier works, both evaluation methods are considered.   

\subsubsection{Evaluation on CSB datasets}
For the evaluation of the CSB datasets, no tolerance region is considered around the ground truth annotation. But, when a neural network is trained on one of the CSB patch datasets it is tested not only on the corresponding test patch subset but also on entire test images of the CSB dataset. 


\section{Experiments and results}
\label{sec:Results and discussion}


\subsection{Experiments and results on CFD dataset}

First, we conduct experiments on the CFD dataset using different neural network architectures. These experiments are done to compare the performances of our neural network architecture proposed in Section \ref{sec:Method} with current state-of-the-art neural networks. For comparison we choose the neural networks proposed by Konig et al. \cite{Konig2021OED} and Liu et al. \cite{Wj2019FPCNet} which show state-of-the-art performances on the CFD dataset according to Table \ref{tab:CFD other works results: 2pixel} in \ref{app:B}. Performances of these methods are reported in the original works, but we also re-implemented them achieving similar performances on the CFD dataset.


Table \ref{tab:Results on CFD} presents the results of these experiments. The performances are evaluated with the 2 pixels of tolerance region around the ground truth annotation as was proposed in \cite{shi2016CFD} and as was explained in Section \ref{subsec:Evaluation on CFD}. Besides that, we show results for the standard $F1$-score evaluation, i.e. with no tolerance region around the ground truth annotation. The table shows the mean and standard deviation based on 5 runs.

The re-implemented method from Liu et al. (Re. Liu et al. in the table) shows just 0.59\% lower performances compared to the performances reported in the original paper (Liu et al. in the table). The difference may be explained by different data augmentation techniques, deep learning frameworks used, and slightly different evaluation approaches (see the original work \cite{Wj2019FPCNet}).
Re-implementation of the neural network proposed by Konig et al. (Re. Konig et al.) achieves 96.44\textsuperscript{±0.01}\% compared to the original results of 96.88\textsuperscript{±0.32}\%. Similarly, the difference in the results can be explained by different evaluation techniques, where adaptable threshold values were used in the original work. 
With this, we conclude that the reimplementation of the neural network architecture from Konig et. al. \cite{Konig2021OED} and Liu et al. \cite{Wj2019FPCNet} was done correctly.

Further, Table \ref{tab:Results on CFD} presents results for the proposed method, indicated as 'ConvNext \& SD \& A'. The method 'ConvNext \& SD' in the table stands for the proposed method without using attention modules in the neural network architecture (refer to Figure \ref{fig:Network scheme}). Finally, the method ConvNext in the table indicates the performance of our proposed method when no attention modules are used in the neural network architecture, and when the stage-wise learning rate decay is not utilized. 
In this set of experiments, for neither of the three of the proposed neural network architectures, inversion of loss function was used.

\begin{figure*}
    \centering
     \includegraphics[width=0.8\linewidth]{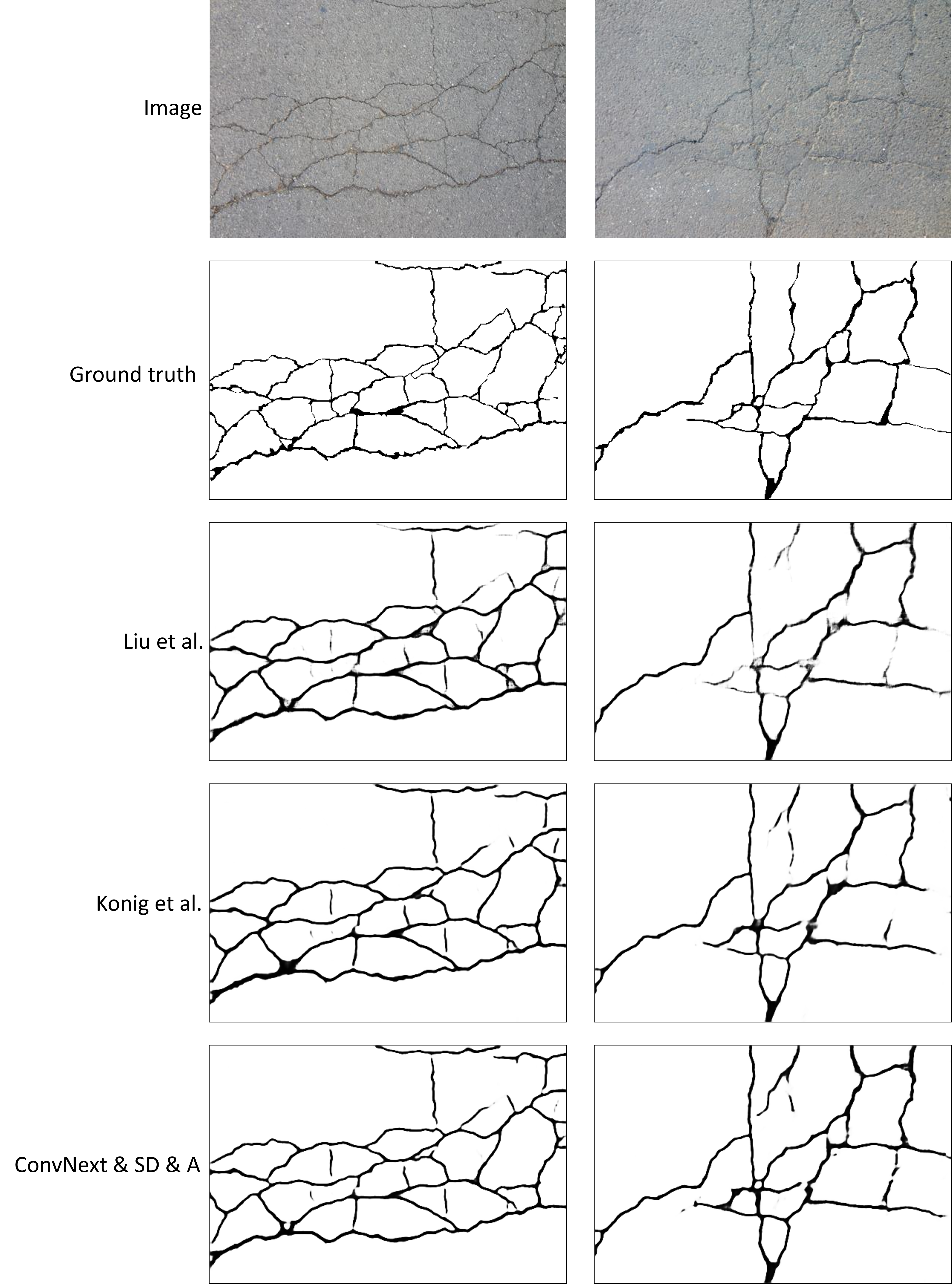}
    \caption{Examples of output segmentation maps of the three methods.}
    \label{fig:CFD output maps}
\end{figure*}

Using Convnext as the neural network encoder with our modified decoder allowed us to increase performance on the CFD dataset by about 0.5\% for evaluations with 2 pixels and 0 pixels tolerance region around the ground truth annotation compared to the method by Konig et al. where a ResNet was used as a neural network encoder. Interestingly, training with stage decay ('ConvNext \& SD' in the table) 
did not give much performance improvement on the CFD dataset.
Using attention modules (ConvNext \& SD \& A) as was explained in Section~\ref{sec:Method}, allows further performance improvements by an insignificant 0.1\% of the $F1$-score, which now reaches 75.35\textsuperscript{±0.14}\% and 97.10\textsuperscript{±0.08}\% with and without considering a tolerance region around the ground truth annotation, respectively.
Figure \ref{fig:CFD output maps} shows some examples of output segmentation maps of two earlier methods and the method we propose in this work. In the given examples, it is possible to see that all methods miss cracks when they are thin. Moreover, Some inconsistency in the annotation can be seen on the right image from Figure \ref{fig:CFD output maps} where similar thin dark lines are marked as cracks in some cases and as background in others. This explains, why all networks make errors on such thin cracks while giving robust segmentation of thick cracks.

\begin{table}
\caption{Results
on the Crack Forest dataset (CFD), \\ where SD stands for `stage decay', A for `attention' and N/A means data not available
\label{tab:Results on CFD}}

\centering
\begin{scriptsize}

\begin{tabular}{l l l l ll l l} \hline 
 & \multicolumn{3}{c}{0 pixels} &  &\multicolumn{3}{c}{2 pixels} \\ \cline{2-4} \cline{6-8}
 Method& $Pr$ & $Re$ & $F1$ &  &$Pr$ & $Re$ & $F1$ \\ \hline 
Liu et al. \cite{Wj2019FPCNet}& N/A & N/A & N/A &  &97.48 & 96.39 & 96.93 \\ 
Konig et al. \cite{Konig2021OED}& N/A & N/A & N/A &  &N/A & N/A & 96.88\textsuperscript{±0.32} \\ 
Re. Liu et al. & 68.59\textsuperscript{±0.53} & 80.81\textsuperscript{±0.43} & 74.20\textsuperscript{±0.13} &  &97.41\textsuperscript{±0.10} & 95.49\textsuperscript{±0.11} & 96.44\textsuperscript{±0.01}\\ 
Re. Konig et al. & 68.26\textsuperscript{±0.75} & 82.12\textsuperscript{±0.65} & 74.54\textsuperscript{±0.24} &  &96.68\textsuperscript{±0.47} & 96.37\textsuperscript{±0.27} & 96.53\textsuperscript{±0.17} \\ 
Convnext & 70.08\textsuperscript{±0.89} & 81.49\textsuperscript{±0.86} & 75.25\textsuperscript{±0.19} &  &97.92\textsuperscript{±0.21} & 96.12\textsuperscript{±0.34} & 97.01\textsuperscript{±0.12} \\ 
Convnext \& SD& 70.50\textsuperscript{±0.96} & 80.65\textsuperscript{±1.31} & 75.22\textsuperscript{±0.14} &  &98.19\textsuperscript{±0.31} & 95.84\textsuperscript{±0.68} & 97.00\textsuperscript{±0.25} \\ 
Convnext \& SD \& A& 70.83\textsuperscript{±0.31} & 80.68\textsuperscript{±0.63} & 75.35\textsuperscript{±0.14} &  &97.93\textsuperscript{±0.17} & 96.29\textsuperscript{±0.22} & 97.10\textsuperscript{±0.08} \\ \hline 

\end{tabular}

\end{scriptsize}
\end{table}

\subsection{Experiments and results on CSB dataset}
Further, we investigate the performances of the neural networks when applied to images of steel bridges, the CSB dataset.
Table  \ref{tab:Results on CSB} presents performances of neural networks that were trained on the CSB 70/30 patch dataset using the sum of Dice loss and BCE loss as a loss function (without inversion). The testing is done on the test subset of the CSB 70/30 patch dataset which has the same crack-to-background patch ratio as in the training subset. This testing is done to understand, how well the neural network is able to perform on the data similar to the one, which was used for training. 
Besides, the neural networks, trained on the CSB 70/30 patch dataset, are tested on entire test images of the CSB dataset and the results are also shown in Table \ref{tab:Results on CSB}. This additional testing is done to estimate the performances on entire images, that can be obtained during real-world automatic inspection of bridges. 
From Table  \ref{tab:Results on CSB}  it can be seen that the neural network proposed by Liu et al. and the neural network proposed by King et al. perform almost equally reaching 72.49\textsuperscript{±1.28}\%  and 71.55\textsuperscript{±1.27}\%  of $F1$-score respectively when tested on the CSB 70/30 patch dataset. Surprisingly, our neural network without attention modules and without stage learning rate decay (ConvNext) does not give much performance gain over baseline methods. But, the addition of the stage decay of the learning rate (ConvNext \& SD) increases the performance of the neural network up to 79.10\textsuperscript{±1.29}\%, when tested on the CSB 70/30 patch dataset. Finally, the addition of the attention module further increases the $F1$-score by a marginal 0.54\%. 

\begin{figure*}
    \centering
    \includegraphics[width=0.8\linewidth]{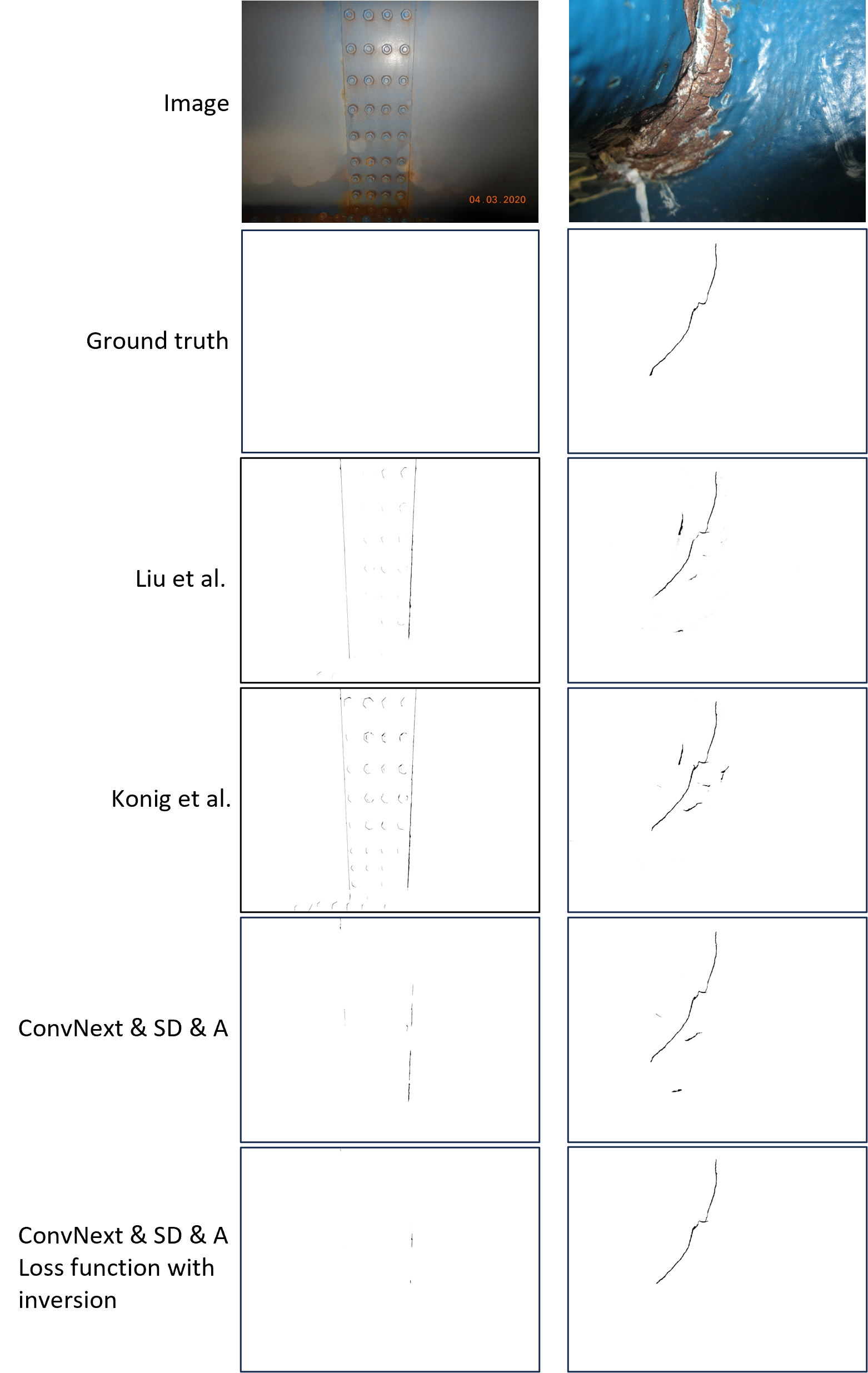}
    \caption{Examples of output segmentation maps of four neural networks  for image without crack (left) and with crack (right)}
    \label{fig:CSB output maps}
\end{figure*}

The right part of Table \ref{tab:Results on CSB} reveals the problem of training a neural network for crack detection using a patch approach, when not all patches from entire images are used. The right part of Table \ref{tab:Results on CSB} shows performances of the neural network trained on train patches of the CSB 70/30 patch dataset, when tested on entire test images of the CSB dataset. While recall remains at a similar level compared to the testing on the CSB 70/30 patch dataset, the precision drops almost by a factor of two for all neural networks. From Eqs.~\! (\ref{eq:Precision}), (\ref{eq:Recall}) it can be seen that low values of precision compared to values of recall indicate a high false positive rate of the neural network. This fact is supported by Figure \ref{fig:CSB output maps}, where output segmentation maps of the neural networks, when applied to two test images from the CSB dataset are shown. The left part of Figure \ref{fig:CSB output maps} shows the output maps for an image without a crack and the right part for an image with a crack. It can be seen from these two examples, that a high false positive rate is a dominant problem in these two images, where crack-like image features such as structure edges, shadows, and corrosion were identified as cracks by most of the neural networks. The same problem we observe for most of the test images of the CSB dataset. The best performance was achieved by the neural network we proposed in this work (ConvNext \& SD \& A) which reaches the 60.29\textsuperscript{±4.00}\% of $F1$-score. Yet the dominant problem is the high false positive rate, which holds the value of precision below 50\% and causes the gap between precision and recall at the level of 25\%. 

The problem of the high false positive rate is caused by two factors: a) the fraction of background patches in the training data (CSB 70/30 patch dataset), which is 30\%,  is much less than the fraction of background patches in real data (CSB dataset), which is 91.3\%; b) the loss function which was used for these experiments, namely the sum of BCE loss and Dice loss, is not sensitive to the mistakes of the neural networks on background patches. 

\begin{table}
\caption{Results of the neural networks trained on CSB 70/30 patch dataset. Testing is done on the CSB 70/30 dataset and the entire test images of the CSB dataset. \\SD stands for `stage decay' and A for `attention'
\label{tab:Results on CSB}}
\centering
\begin{scriptsize}

\begin{tabular}{l l l l ll l l} \hline
 & \multicolumn{3}{c}{Test CSB 70/30 patch dataset} &  &\multicolumn{3}{c}{Test CSB dataset} \\ \cline{2-4} \cline{6-8}
Method & $Pr$ & $Re$ & $F1$ &  &$Pr$ & $Re$ & $F1$ \\ \hline
Re. Liu et al. & 70.06\textsuperscript{±2.38}& 74.10\textsuperscript{±1.94}& 72.49\textsuperscript{±1.28}&  &32.93\textsuperscript{±3.93}& 77.08\textsuperscript{±1.48}& 46.01\textsuperscript{±3.87}\\ 
Re. Konig et al. & 71.98\textsuperscript{±3.16}& 71.03\textsuperscript{±2.26}& 71.55\textsuperscript{±1.27}&  &33.22\textsuperscript{±2.24}& 73.15\textsuperscript{±1.73}& 45.61\textsuperscript{±1.93}\\ 
ConvNext & 70.87\textsuperscript{±2.71}& 75.27\textsuperscript{±1.52}& 72.44\textsuperscript{±2.12}&  &31.77\textsuperscript{±3.99}& 78.12\textsuperscript{±1.01}& 45.01\textsuperscript{±4.20}\\ 
 ConvNext \& SD& 81.46\textsuperscript{±1.43}& 77.65\textsuperscript{±2.03}& 79.10\textsuperscript{±1.29}&  &49.51\textsuperscript{±3.67}& 76.378\textsuperscript{±3.32}& 59.88\textsuperscript{±1.89}\\ 
ConvNext \& SD \&  A& 81.10\textsuperscript{±0.66}& 78.24\textsuperscript{±2.96}& 79.63\textsuperscript{±1.86}&  &49.43\textsuperscript{±5.60}& 74.492\textsuperscript{±7.20}& 60.29\textsuperscript{±4.00}\\ 
\hline
\end{tabular}
\end{scriptsize}
\end{table}

\subsection{Effect of background patches}

To solve the problem of the high false positive rate of the neural network when tested on entire images of the CSB dataset, we conduct another set of experiments exploring the effect of the number of background patches used for training and the effect of the proposed loss function with inversion. 
This set of experiments is done with the neural network which showed the best performance in the described above experiments - the neural network we proposed in Section \ref{sec:Method} of this paper (ConvNext \& SD \& A).  
First, we experiment with different CSB patch dataset compositions to find how close the ratio of the patches in the train dataset can be to 6.45\%/93.55\%, which is the crack-to-background patches ratio in the full CSB dataset, and yet, not causing the "all-white segmentation maps" trap. When we reach the performance boundary, we employ the loss function with inversion that allows to have more background patches in the train dataset while still avoiding the "all white segmentation maps" problem.

Table \ref{tab:Results on CSB:different patch composition} shows results of experiments with different crack patch to background patch ratios and different loss functions. As these experiments are computationally demanding, due to the higher number of training patches, results in Table \ref{tab:Results on CSB:different patch composition} are based on a single run. It can be seen that increasing the number of background patches so that they provide 70\% of all patches, allowed to increase recall by 5\% and precision by almost 10\%. These results show that showing more background patches allows a neural network to better distinguish crack-like image features from cracks, making fewer false positive errors, but also increases its ability to recognize the cracks. 

However, further increase in the number of background patches used for neural network training has a negative effect. 
Training on the 10/90 patch dataset, showed that if the fraction of background patches in the patch dataset reaches 90\%, the neural network gets into the "all-white segmentation maps" trap, where the neural network learns to always predict the background on an image. The neural network that gets in this trap has precision equal to 1, recall equal to 0, and an $F1$-score equal to 0 according to Eqs.~\!\! (\ref{eq:Precision}), (\ref{eq:Recall}) and (\ref{eq:F1-score}), because then the  FP = 0 and TP = 0,  on each image.  The "all-white segmentation map" is partially caused by the training data imbalance issue. While trained on the CSB 10/90 patch dataset, a neural network learns to ignore any crack. On 90\% of the patches, it would still have perfect performance because 90\% of the patches are background patches.  To avoid this local minimum in the optimization space, first, we applied a Tversky loss function \cite{salehi2017tversky}, which weighs unequally false positives and false negatives in the DICE loss. Specifically, we set weight for false positives $\alpha$=0.01 and weight for false negatives $\beta$ = 0.99. However, it did not solve the problem. The reason why the Tversky loss function did not help to solve the problem is that it is still unable to give an accurate estimate of the error, that the neural network makes on background patches. 

As was described in Section \ref{sec:Method}, the inversion of the loss function helps to avoid the "all-white segmentation maps" problem by providing a better error estimate on background patches. Using this modified loss function we were able to further increase the overall performance of the neural networks on the CSB dataset, by reducing the gap between precision and recall to 5\% and increasing the $F1$ score to 70.79\%. Finally, we train the neural network on the CSB dataset, which contains 3 950 (8.7\%) of crack patches and 57 276 (91.3)\% of background patches. Indeed, training on this data with an ordinary sum of BCE loss and dice loss causes the "all-white segmentation map" problem. However, the loss function with inversion allowed to avoid this trap and train a neural network that reaches an $F1$-score of 71.99\%.
This is achieved thanks to the sensitivity of the loss function with inversion to the neural network errors on background patches. The normal sum of the BCE loss and Dice loss would be a similar loss to the neural network errors, which give a high amount of FP or a low amount of FP on the background patch. But the loss function with inversion assesses the error of the background segmentor when a background patch is encountered during training, thus being more sensitive to the errors on the background patches. 

\begin{table}
\caption{Results of the proposed neural network 
on the Cracks in Steel Bridges dataset (CSB) with different loss functions and patch dataset composition 
\label{tab:Results on CSB:different patch composition}}
\centering
\begin{scriptsize}
\begin{tabular}{ l  ll  l  l } \hline

Train dataset &  Loss function&$Pr$ & $Re$ & $F1$ \\ \hline

CSB 70/30 patch dataset&  Dice+BCE&49.43\textsuperscript{±5.60}& 74.49\textsuperscript{±7.20}& 60.29\textsuperscript{±4.00}\\ 

CSB 30/70 patch dataset&  Dice+BCE&58.02 & 79.27 & 67.00 \\ 

CSB 10/90 patch dataset&  Dice+BCE&1 & 0 & 0 \\ 
 CSB 10/90 patch dataset &  Tversky loss: a=0.01 b=0.99&1& 0&0\\ 
 CSB 10/90 patch dataset &  Dice+BCE with inversion&68.498 & 73.24 &70.79 \\ 
 CSB dataset&  Dice+BCE&1& 0&0\\ 

CSB dataset&  Dice+BCE with inversion&68.73& 75.59& 71.99\\ 
\hline
\end{tabular}
\end{scriptsize}
\end{table}


\subsection{Effect of annotation errors}

As explained in Section \ref{sec:Datasets}, to produce ground truth segmentation for the CSB datasets, a semi-automatic tool was used which significantly reduces annotation time, but also reduces annotation accuracy. The annotation for the CSB patches dataset was manually corrected, to reduce the impact of the annotation errors on neural network performance. We make an additional study to investigate the effect of annotation error by training the proposed neural network on the dataset that did not undergo manual correction. Comparing it with a neural network trained on a corrected dataset allowed us to estimate the performance drop due to the annotation errors induced by the automatic segmentation tool proposed in \cite{CP1}. Here, as a loss function, we use the sum of BCE and Dice loss (Eq. \ref{eq:Dice + BCE}) without applying the inversion as was proposed in Section \ref{sec:Method}.

The results are shown in Table \ref{tab:Results on uncorrected dataset}, where we compare the performance of our proposed method when it is trained on the CSB 70/30 dataset with manually corrected annotation and without manual correction. We can see that the annotation errors cause just a 1\% drop in $F1$-score when tested on the CSB 70/30 patch dataset and less than 2\% when tested on the CSB dataset. With this, we conclude that the semi-automatic crack annotation tool \cite{github_segmentation_tool} (summarized in~\ref{app:A}) which was used in this study causes a minor effect on the performance of the neural network. 
\begin{table}
\caption{Results of the neural networks trained on CSB 70/30 dataset which was not manually corrected. Testing is done on the CSB 70/30 dataset and the full CSB dataset
\label{tab:Results on uncorrected dataset}}
\centering
\begin{scriptsize}
\begin{tabular}{ l  l  l  l  ll  l  l } \hline
 & \multicolumn{3}{c}{CSB 70/30 patch dataset}&  &\multicolumn{3}{c}{CSB dataset}\\ \cline{2-4} \cline{6-8}
 Dataset correction& $Pr$& $Re$& $F1$&  &$Pr$& $Re$&$F1$\\ \hline

Corrected & 81.10\textsuperscript{±0.66} & 78.24\textsuperscript{±2.96} & 79.63\textsuperscript{±1.86} &  &49.43\textsuperscript{±5.60} & 74.49\textsuperscript{±7.20} & 60.29\textsuperscript{±4.00} \\ 

Not corrected & 76.66\textsuperscript{±0.42} & 86.28\textsuperscript{±0.51} & 78.97\textsuperscript{±0.45} &  &46.80\textsuperscript{±4.19} & 78.84\textsuperscript{±2.23} & 58.64\textsuperscript{±3.54} \\ 

\hline
\end{tabular}
\end{scriptsize}
\end{table}

\subsection{Effect of patch size}

It is known that reducing the patch size reduces the computational effort. We conduct additional experiments to show how the patch size affects the performance of the neural network for crack segmentation. For this, we use the CSB 70/30 patch dataset, where each patch is reduced, first to the size of 384x384 pixels and then to the size of 128x128 pixels. To reduce the patch size, we crop a region of the initial patch in such a way that if a crack is present, it remains in the reduced patch. The reduced patches and the implementation of the code for patch size reduction are also provided, see \cite{CSB_dataset}

\begin{figure}
  \centering
  \subfigure[]{
    \includegraphics[width=0.4\textwidth]{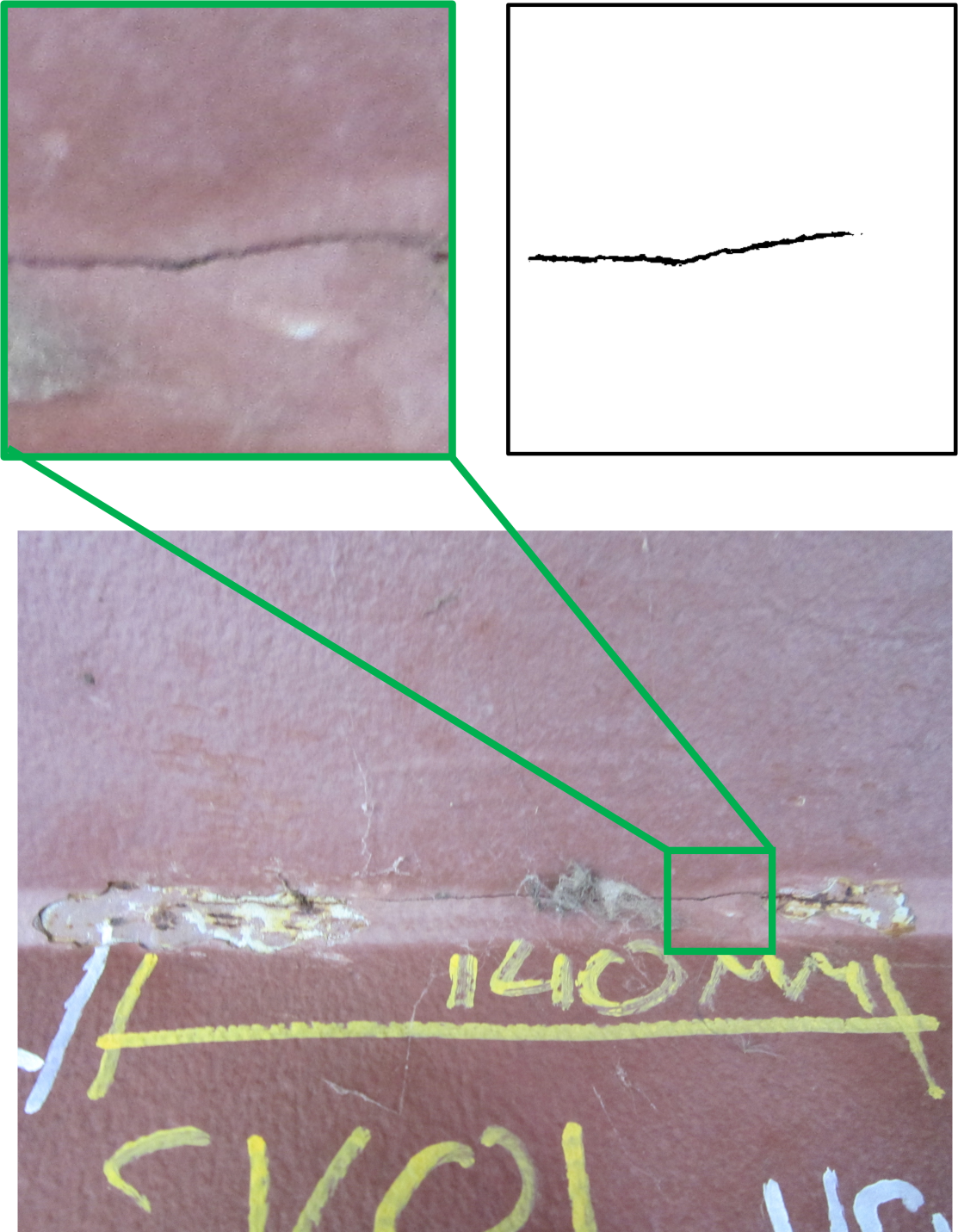}
    \label{fig:global context a}
  }
  \subfigure[]{
    \includegraphics[width=0.4\textwidth]{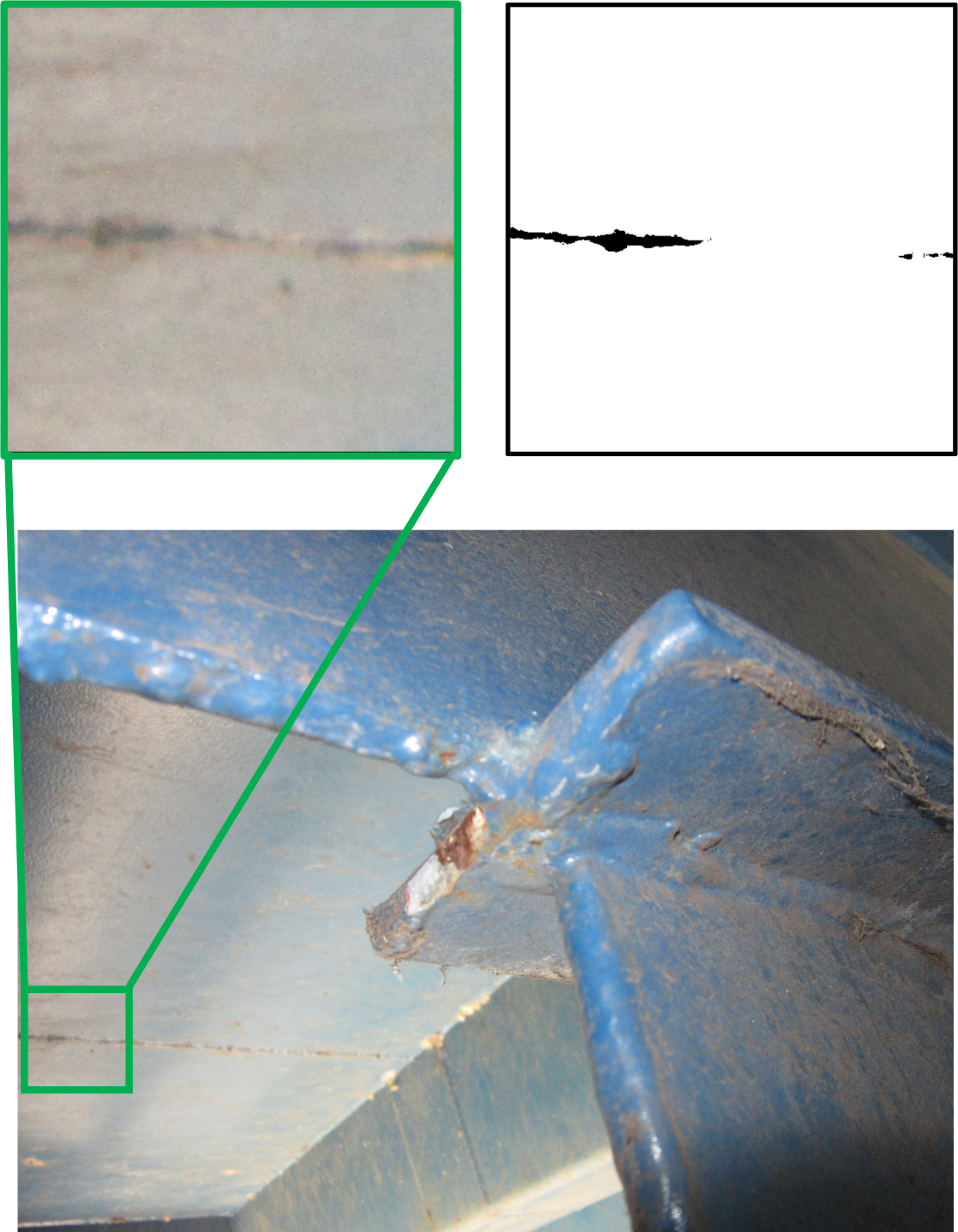}
    \label{fig:global context b}
  }
  \caption{Visualisation of the need for including the global context in order to distinguish crack from crack-like features of an image (e.g. structural joint). We depict examples with an entire image, a patch, and the corresponding neural network segmentation results: a) example with a crack; b) example with a `crack-like image feature'}
  \label{fig:global context}
\end{figure}

In Table \ref{tab:Results on CSB:different patch size} can be seen that when the neural network is trained on patches with smaller sizes and tested on entire images, the $F1$-score significantly drops. It also can be noted, that while the precision also decreases significantly to the value of 15\%, the recall value increases to 83.04\%. This means that the neural networks miss fewer cracks but result in a significant increase in the false positive rate. We can conclude that when a neural network is trained on smaller patches, it sees less of a global context and it is unable to distinguish crack-like image features from cracks. Figure \ref{fig:global context} demonstrates this effect. Figure \ref{fig:global context} (a) shows an entire image with a crack, a crack patch, and a segmentation map produced by a neural network for this patch. In figure \ref{fig:global context} (b) an entire image without a crack is shown, and a patch is taken which captures a crack-like image feature. While looking only at the patches in Figure \ref{fig:global context}, there is not enough information even for a human expert to confidently conclude if the patches contain a crack or a crack-like image feature. However, when a "global context" from the entire image is observable, it becomes easier to distinguish the crack from the crack-like image feature. For example, it can be seen that the patch on figure \ref{fig:global context} (a) is taken from a weld location where the crack is most probable to occur. Furthermore, the shape of the dark line and the corrosion at its tips allow to conclude that this is a crack. In contrast, the part of the dark line captured by a patch in Figure \ref{fig:global context} (b) is taken from a flat surface of a structure, and the fact that the dark line is long and straight helps to conclude that this dark line is a joint of a structure rather than a crack. Therefore, we conclude that the global context is important, while cracks are being recognized on an image and our further research will be aimed at investigating the ways to incorporate the global image information into a crack detection and segmentation framework.

\begin{table}
\caption{Performances of the proposed neural network when trained on CSB 70/30 patch datasets with different patch sizes 
\label{tab:Results on CSB:different patch size}}
\centering
\begin{scriptsize}
\begin{tabular}{ l  l  l  l  ll  l  l } \hline
 & \multicolumn{3}{c}{Test CSB 70/30 patch dataset}&  &\multicolumn{3}{c}{Test CSB dataset}\\ \cline{2-4} \cline{6-8}
 Patch size& $Pr$& $Re$& $F1$&  &$Pr$& $Re$&$F1$\\ \hline

512& 81.10\textsuperscript{±0.66} & 78.24\textsuperscript{±2.96} & 79.63\textsuperscript{±1.86} &  &49.43\textsuperscript{±5.60} & 74.49\textsuperscript{±7.20} & 60.29\textsuperscript{±4.00} \\ 

384& 80.83\textsuperscript{±1.20} & 78.78\textsuperscript{±1.94} & 79.77\textsuperscript{±1.01} &  &44.79\textsuperscript{±3.49} & 80.06\textsuperscript{±1.57} & 57.34\textsuperscript{±2.67} \\ 

128& 61.00\textsuperscript{±3.35} & 82.25\textsuperscript{±1.40} & 70.77\textsuperscript{±2.32} &  &17.87\textsuperscript{±2.01} & 83.04\textsuperscript{±2.11} & 29.35\textsuperscript{±2.79} \\ \hline

\end{tabular}
\end{scriptsize}
\end{table}


\section*{Acknowledgements}
The authors would like to thank the Dutch bridge infrastructure owners "ProRail" and "Rijkswaterstaat", and "Nebest" engineering company for their support. 
The research is primarily funded by the Eindhoven Artificial Intelligence Systems Institute, and partly by the Dutch Foundation of Science NWO (Geometric Learning for Image Analysis, VI.C 202-031).
\section{Conclusion}
\label{sec:Conclusion}

We propose a neural network architecture for segmenting cracks in real steel bridge structure, which combines a novel ConvNext neural network with the previously introduced encoder-decoder neural network for crack segmentation.
Moreover, we introduced a novel and challenging dataset for the segmentation of fatigue cracks in images of steel bridges, which has been annotated with a semi-automatic segmentation tool. 

The main findings are:

(1) When image patches are used to train a neural network to segment cracks on entire large images, the use of a substantial amount of background patches allows for avoiding a high false positive rate. In other words, the composition of the patch dataset on which a neural network is trained has a significant effect on its performance;

(2) The common loss functions such as binary cross entropy loss and dice loss lack sensitivity in the estimate of neural network mistakes when used with background patches;

(3) We introduced the concept of loss function inversion, which solves the aforementioned problem and allows for a more efficient use of background patches during training, allowing us to achieve 71.99\% of F1-score when applied to the proposed dataset;

(4) Patch size has a significant effect on neural network performance when tested on entire images, meaning that the global information contained in the entire image contributes to distinguish crack-like features from actual cracks. 

{\small
\bibliographystyle{elsarticle-num}
\bibliography{references} 
}

\appendix
\section{Semi-automatic Segmentation Tool for Annotation in our Dataset of Cracks in Steel Bridges\label{app:A}}

In this section, we briefly summarize our geometric semi-automatic annotation tool for accurate pixel-wise annotation of cracks on images, which allowed a considerable speed-up in obtaining accurate annotations compared to a fully manual approach  \cite{CP1}.  

The tool relies on user input, which indicates the location of two endpoints of a crack on an image. An optimal path algorithm is used to find a crack path on the image between the selected endpoints. To get a more accurate crack path, the optimal path algorithm is applied to the image transformed into so-called orientation scores. Afterward, the crack edges around the obtained crack path are identified, allowing the crack segmentation. 
The implementation of the developed tool with GUI is available at \cite{github_segmentation_tool}. 

\subsection{Construction of orientation score}
As was mentioned, the images of cracks in steel bridges contain lots of crack-like image features. These crack-like image features may intersect a crack and mislead the optimal path algorithm. 
To diminish the effect of crack-like features the optimal path algorithm is applied in a so-called orientation score \cite{duitsAMS} of the image which is obtained through a multi-orientation wavelet transform. This transform is used to separate the image line features with different orientations, allowing better tracking of the crack path. Moreover, tracking of a crack in an orientation score allows to smoothly follow the crack in the regions where its visibility is minimal
 \cite{CP1,vdBerg2024}. 
To transform an image into an orientation score the cake wavelets are applied as follows. 

\begin{figure}
  \centering
  \subfigure[]{
    \includegraphics[width=0.4\textwidth]{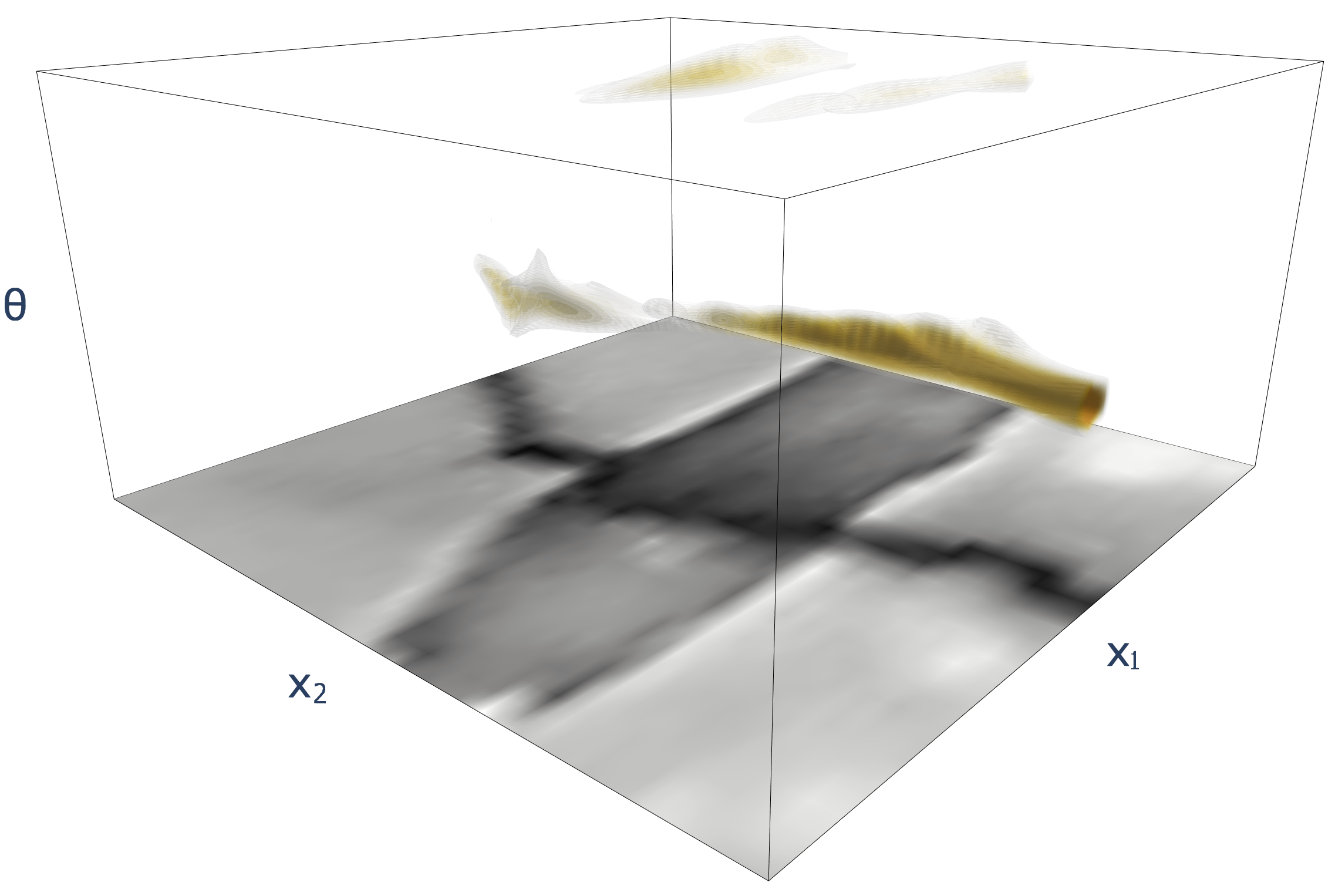}
    \label{fig:orientation scores of crack}
  }
  \subfigure[]{
    \includegraphics[width=0.4\textwidth]{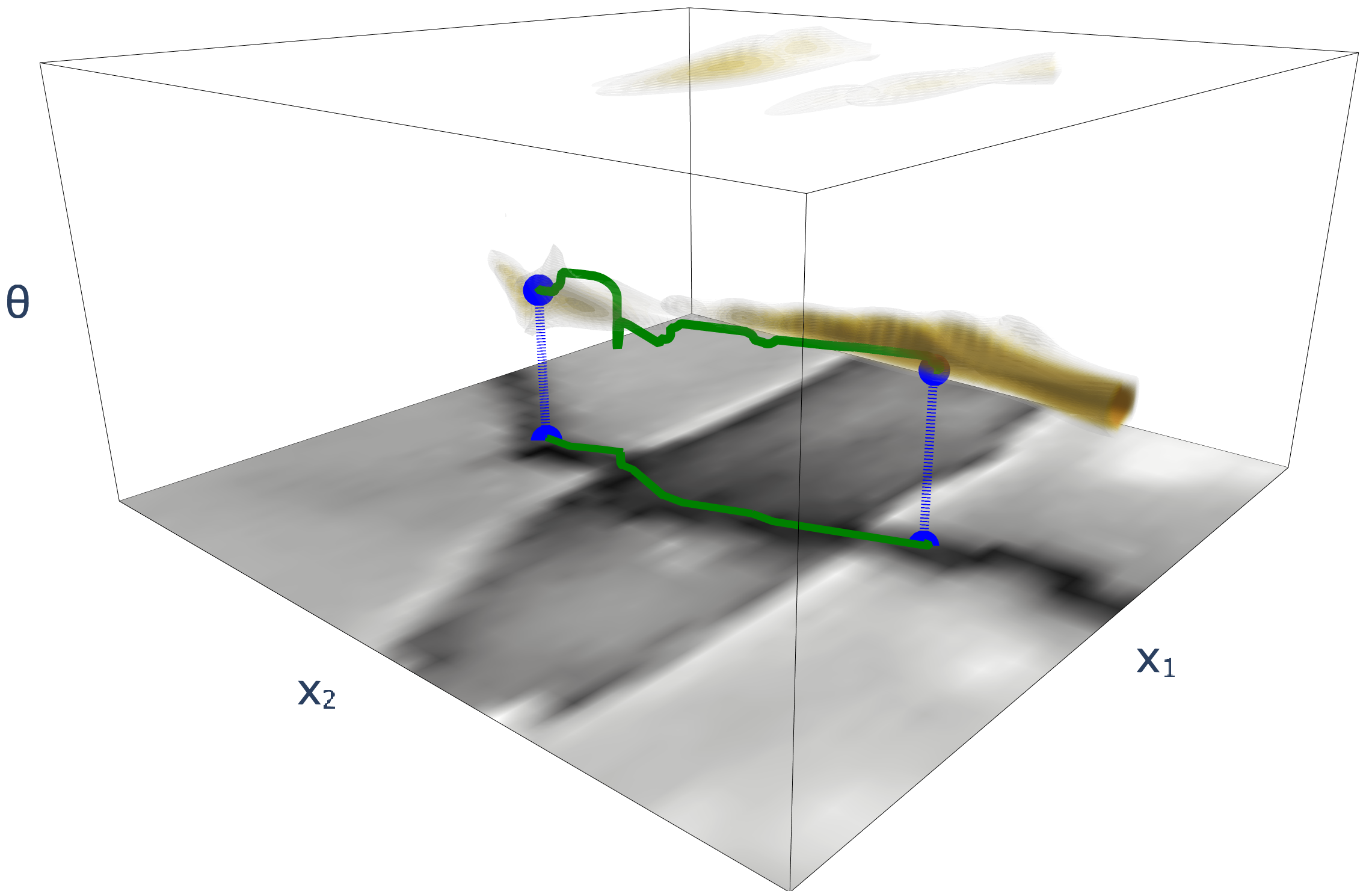}
    \label{fig:orientation scores with track}
  }
  \caption{a) Orientation score; b) An example of a geodesic (the shortest path $ \gamma$ from Eq. \ref{dist}) in the orientation score and its projection onto $ \R^2$.}
  \label{fig:orientation score of crack + orientation score of crack with track}
\end{figure}

A grayscale image can be considered as a (square-integrable) function $f:\R^2 \to \R$ that maps a position $\mathbf{x} \in \R^2$ to a grayscale value $f(\mathbf{x})$. Similarly, the orientation score $U_f$ is represented by the function 
\mbox{$U_{f}:\mathbb{M}_2\rightarrow \mathbb{C}$}, where \mbox{$ \mathbb{M}_2 = \mathbb{ R}^{2}\times S^1$} and \mbox{$S^1:=\left\{\left.\mathbf{n}(\theta)=(\cos \theta,\sin \theta)\;\right|\; \theta \in [0,2\pi)\right\}$}. 
The used cake wavelets $\psi$ are complex-valued (their real part detects lines whereas their imaginary part detects edges). 
The orientation score $U_f$ of an image $f$ is 
given by:

{\scriptsize
\begin{equation}\label{Uf}
U_f(\mathbf{x},\theta) = \int_{\mathbb{R}^2} \overline{\psi \left( R_\theta^{-1}(\mathbf{y} - \mathbf{x}) \right)} f(y)\, \mathrm{d}\mathbf{y}, \textrm{ for all } \mathbf{x} \in \R^2, \theta \in [0,2\pi),
\end{equation}} 

where $ \psi $ is the complex-valued cake wavelet~\cite{PhDThesisRDuits,duitsAMS}  aligned with an a priori axis (say the vertical axis $\theta=0$), and the rotation matrix $ R_\theta$ rotates this cake wavelet counterclockwise with the required angle $\theta$ and is defined by:
{\small $R_\theta = 
\begin{pmatrix}
\cos\theta & -\sin\theta\\
\sin\theta & \cos\theta
\end{pmatrix}
$}. Figure \ref{fig:orientation scores of crack} shows a zoomed-in grey-scale image of a crack with crack-like features and the orientation score of this image. The crack in Figure \ref{fig:orientation scores of crack} is the darkest line oriented parallel to $\mathbf{x_2}$ axis and crack-like features are oriented perpendicular to the crack. It can be seen from this example how the crack and the crack-like features that intersect the crack are separated in orientation score. This separation has a positive effect on the optimal path retrieval algorithm.  

\subsection{Shortest paths (geodesics) in the orientation score} 
An optimal path algorithm is applied in the created orientation score, which finds an optimal path, or geodesics, between the manually selected end-points. The geodesics represent the crack path in the orientation scores. 
The geodesic can be represented by a parameterized curve $\gamma(t)=(\mathbf{x}(t),\mathbf{n}(t))$ in the orientation scores, the length of which is defined as the Riemannian distance between the chosen endpoints $\mathbf{p} = (\mathbf{x}_0,\mathbf{n}_0)$ and $ \mathbf{q}=(\mathbf{x}_1,\mathbf{n}_1)$. Here, $\mathbf{p}$ and $\mathbf{q}$ are points in the lifted space of positions and orientations $ \mathbb{M}_2:=\mathbb{R}^2\times S^1$ and $\mathbf{n}(t):= \begin{pmatrix}
    \cos{\theta(t)} \\ 
    \sin{\theta(t)} 
\end{pmatrix}$. The asymmetric version of the Riemannian distance, that is used, is defined by:

\begin{equation} \label{dist}
d_G(\mathbf{p},\mathbf{q}) = \inf_{\substack{\gamma(\cdot)=
\left(\mathbf{x}(\cdot), \mathbf{n}(\cdot)\right) \in \Gamma\\ \gamma(0)=\mathbf{p}, \gamma(1)=\mathbf{q} \\ \dot{\mathbf{x}}(\cdot) \cdot \mathbf{n}(\cdot) \geq 0}} \int_{0}^{1} \sqrt{G_{\gamma(t)} (\dot{\gamma}(t),\dot{\gamma}(t))} \mathrm{d}t
\end{equation}
where the space of curves $\Gamma$ (are all piecewise continuously differentiable curves $\gamma:[0,1]\rightarrow \mathbb{M}_2$) over which we optimize,
and with the Riemannian metric given by {\small
\begin{equation}\label{G} 
G_\mathbf{p}(\dot{\mathbf{p}},\dot{\mathbf{p}}) \!=\! C^2(\mathbf{p})\!\left(\! \xi^2|\dot{\mathbf{x}}\cdot \mathbf{n}|^2\! + \! \frac{\xi^2}{\zeta^2} \left(\lVert \dot{\mathbf{x}}\rVert^2\!\!-\!|\dot{\mathbf{x}}\cdot \mathbf{n}|^2\right) \!+\! \lVert \dot{\mathbf{n}} \rVert^2 \!+\! 
\lambda 
{\scriptsize
\frac{\max\limits_{\|\dot{\mathbf{q}}\|=1} \left| HU|_\mathbf{p}(\dot{\mathbf{p}},\dot{\mathbf{q}})\right|^2}{\max\limits_{\substack{\|\dot{\mathbf{q}}\|=1,\\\|\dot{\mathbf{p}}\|=1}} \left| HU|_\mathbf{p}(\dot{\mathbf{p}},\dot{\mathbf{q}}) \right|^2}} \!\right)
\end{equation}}with $\mathbf{p}=(\mathbf{x},\mathbf{n}) \in \mathbb{M}_2$ being a position and orientation, and $\dot{\mathbf{p}}=(\dot{\mathbf{x}},\dot{\mathbf{n}})$ a velocity attached to the point $\mathbf{p}$, and 
where $ C^2(\mathbf{p})$ is the output of the application of the multi-scale crossing/edge preserving line filter to the orientation score, for details see \cite[App.D]{vdBerg2024}, \cite{Hannink2014}, and acts as a cost function. The Hessian of the orientation score is denoted by $ HU$. 
Parameter $\xi>0$ influences the stiffness or curvature of the geodesics. Parameter $0<\zeta \ll 1$ puts a high cost on $ \xi^2/ \zeta^2$ on sideward motion relative to the cost $ \xi^2$ for forward motion. Parameter $\lambda>0$ regulates the influence of the data-driven term relying on the Hessian of the orientation scores.

As a numerical implementation of the given above mathematical formulation, an anisotropic fast marching algorithm is used to compute the distance map $d_{G}(\mathbf{p},\cdot)$ from one of the crack end-points $\mathbf{p}$. The computed distance map is used to backtrack the geodesics from the other crack end-point $\mathbf{q}$, using the steepest descent method \cite{Duits2018_27}. 

To obtain the crack path in the initial 2D image, the obtained geodesics from the orientation score is simply projected back on the initial image along $\theta$ axis of the orientation score. Figure \ref{fig:orientation scores of crack} demonstrates a geodesic in the orientation score and its spatial projection on the initial image. 

\subsection{Crack width detection}
After the crack path on an image is identified the final step of the semi-automatic tool determines the width of a crack at each location along its path. To do so, line filtering is done using a Gaussian first-order derivative filter, oriented in a direction, perpendicular to the local crack track orientation. This filtering allows to extraction of edges of a crack along its path. Afterwards, an anisotropic fast marching algorithm is applied to the image with filtered crack edges (without constructing orientation scores) in order to identify the crack edges along the crack track. This completes the semi-automatic segmentation of a crack.



\section{Previous work on crack segmentation \label{app:B}}
In this section, we provide some background information on existing datasets for crack segmentation (Table \ref{tab:Datasets list}) and algorithms developed earlier for segmentation of cracks in the CFD dataset (Tables \ref{tab:CFD other works results: 0pixel}, \ref{tab:CFD other works results: 2pixel}, \ref{tab:CFD other works results: 5pixel}).

\begin{table}
\caption{Datasets for segmentation of cracks in images. Table is based on \cite{li2022review,zhou2023deep}
\label{tab:Datasets list}}
\centering
\begin{scriptsize}
\begin{tabular}{ l  l  l  l }\hline

Name & Availability & Objects/Materials & Num. of images (size) \\ \hline

CRACK500 \cite{yang2019FPNHB}& Public & Pavement & 500 (2000 x 1500) \\

GAPs 10m \cite{stricker2021road}& Public & Pavement & 20 (5030 x 11505) \\

CrackTree260 \cite{zou2018deepcrack}& Public & Pavement & 260 (various sizes) \\

Stone331 \cite{zou2018deepcrack}& Public & Stone & 331 (512 x 512) \\

CRKWH100 \cite{zou2018deepcrack}& Public & Pavement & 100 (512 x 512) \\

CrackLS315 \cite{zou2018deepcrack}& Public & Pavement & 315 (512 x 512) \\

CrackForest \cite{shi2016CFD}& Public & Pavement & 118 (320 x 480) \\

DeepCrack \cite{liu2019deepcrack}& Public & Concrete + pavement & 537 (544 x 384) \\

Yandg. CD \cite{yang2018automatic}& Public & Concrete + pavement & 800 \\

CrackIT \cite{oliveira2014crackit}& Public & Pavement & 84 (1536 x 2048) \\

Aigle-RN \cite{amhaz2016AiglRN}& Public & Pavement & 38(991x462) \\

EdmCrack600 \cite{mei2020EdmCrack600} & Public & Pavement & 600 (1920x1080) \\

FIND & Public & Bridge deck, Pavement & 2500 (256x256) \\

IPC-SHM 2020 \cite{bao2021SimilarDataset}& Not public & Steel bridge & 120 (up to 5152 x 3864) \\

\textbf{CSB} (ours) \cite{CSB_dataset} & Public & Steel bridge & 755 (up to 4608 x 3456) \\ \hline

\end{tabular}
\end{scriptsize}
\end{table}

\begin{table}

\centering
\caption{List of publications with results on CFD dataset, using 0-pixel tolerance region around the ground truth annotation for evaluation
\label{tab:CFD other works results: 0pixel}}
\begin{scriptsize}
\begin{tabular}{ l  l  l  l } \hline

Method & $Pr$ & $Re$ & $F1$ \\ \hline

Ai et al. \cite{ai2018automatic}& 47.1 & 75.7 & 56.7 \\

Yang et al. \cite{yang2019feature}&  &  & 70.5 \\

Chen et al. \cite{chen2021HACNet}& 67.8 & 74.7 & 71.0 \\

Augustauskas et al. \cite{augustauskas2020improved}&  &  & 71.21 \\

Rill-Garcia et al. \cite{rill2022pixel}&  &  & 71.77 \\

Qu et al. \cite{qu2021crack}&  &  & 76.3 \\

Fan et al. \cite{fan2023pavement} &  &  & 78.3 \\

Zhong et al. \cite{zhong2022multi}& 78.35  & 80.52 & 79.42 \\

Chu et al. \cite{chu2022tiny}&  &  & 87.06 \\ \hline

\end{tabular}
\end{scriptsize}
\end{table}

\begin{table}
\centering
\caption{List of publications with results on CFD dataset, using 2-pixel tolerance region around the ground truth annotation for evaluation
\label{tab:CFD other works results: 2pixel}}
\begin{scriptsize}
\begin{tabular}{ l  l  l  l } \hline

Method & $Pr$ & $Re$ & $F1$ \\ \hline

Ai et al. \cite{ai2018automatic}& 90.7 & 84.6 & 87 \\

Fan et al. \cite{fan2018automatic}& 91.19 & 94.81 & 92.44 \\

Konig et al. \cite{konig2019segmentation}&  &  & 94.92 \\

Fan et al. \cite{fan2020ensemble}& 95.52 & 95.21 & 95.33 \\

Lau et al. \cite{lau2020automated}& 97.02 & 94.32 & 95.55 \\

Fan et al. \cite{fan2022use}& 96.21 & 95.12 & 95.63 \\

Inoue at al. \cite{inoue2019deployment} &   &   & 95.7 \\

Qiao et al. \cite{qiao2021automatic}& 97.29 & 94.56 & 95.90 \\

Al-Huda et al. \cite{al2023hybrid}& 97.9 & 94.1 & 960 \\

Ong et al. \cite{ong2023feature}& 97.89 & 94.63 & 96.19 \\

Li et al. \cite{li2021novel}& 97.51 & 95.95 & 96.72 \\

Liu et al. (FPCNet) \cite{Wj2019FPCNet}& 97.48 & 96.39 & 96.93 \\

Konig et al. \cite{Konig2021OED}&  &  & 96.97\textsuperscript{±0.23} \\ \hline

\end{tabular}
\end{scriptsize}
\end{table}

\begin{table}

\centering
\caption{List of publications with results on CFD dataset, using 5-pixel tolerance region around the ground truth annotation for evaluation.
\label{tab:CFD other works results: 5pixel}}
\begin{scriptsize}
\begin{tabular}{ l  l  l  l } \hline

Method & $Pr$ & $Re$ & $F1$ \\ \hline

Jenkins at al. \cite{jenkins2018deep}& 92.46 & 82.82 & 87.83 \\

Escolana et al. \cite{escalona2019fully}& 97.31 \textsuperscript{±0.28} & 94.28 \textsuperscript{±0.51}  & 95.75 \textsuperscript{±0.22} \\ \hline

\end{tabular}
\end{scriptsize}
\end{table}





\end{document}